\documentclass[10pt,twocolumn,letterpaper]{article}

\usepackage[pagenumbers]{cvpr} 
\usepackage{multirow}

\usepackage{cuted}
\usepackage{capt-of}
\usepackage{float}
\usepackage{caption}

\definecolor{cvprblue}{rgb}{0.21,0.49,0.74}
\usepackage[pagebackref,breaklinks,colorlinks,allcolors=cvprblue]{hyperref}

\def\paperID{10684} 
\def\confName{CVPR}
\def\confYear{2025}

\title{Hero-SR: One-Step Diffusion for Super-Resolution with \\ Human Perception Priors }

\author{
    Jiangang Wang$^{1,2}$, Qingnan Fan$^{2\dag}$, Qi Zhang$^{2}$, \\ Haigen Liu$^{1,2}$, 
    Yuhang Yu$^{2}$, Jinwei Chen$^{2}$, Wenqi Ren$^{1\dag}$ \\
    $^{1}$School of Cyber Science and Technology, Shenzhen Campus of Sun Yat-sen University \\
    $^{2}$vivo Mobile Communication Co. Ltd \\
    {\tt\small \{wangjg33,liuhg6\}@mail2.sysu.edu.cn, } 
    {\tt\small \{fqnchina,nwpuqzhang\}@gmail.com} \\ 
    {\tt\small yuyuhang@vivo.com, chenjinwei\_1987@126.com, } 
    {\tt\small renwq3@mail.sysu.edu.cn} \\
    {\small Project Page: \url{https://github.com/W-JG/Hero-SR}}
}

\begin{document}

\maketitle

\setlength\stripsep{-15pt}
\begin{strip}
    \centering
\includegraphics[width=1.0\linewidth]{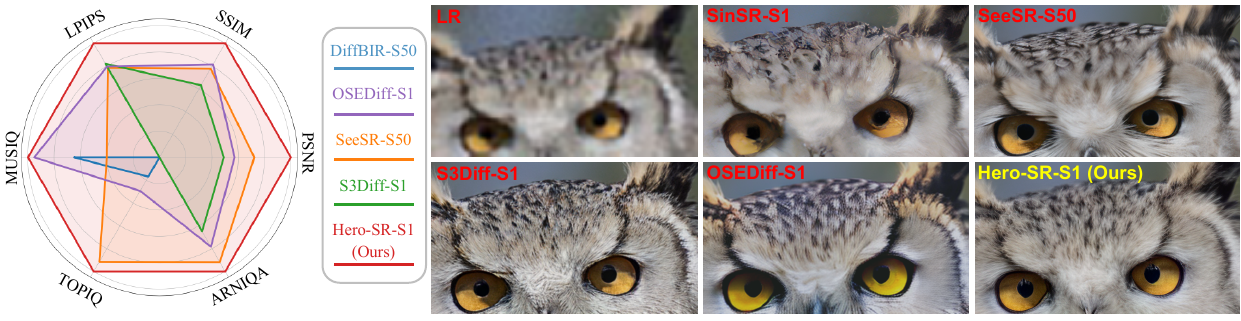}
    \captionof{figure}{Performance and Visual Comparison. (1) Performance Comparison: Compared to one-step and multi-step methods, Hero-SR achieves superior performance with just a single diffusion step. Tested on the DRealSR benchmark, all metrics are normalized using min-max scaling, with `S' denoting the number of diffusion steps. (2) Visual Comparison: Hero-SR restores more realistic textures and aligns better with human perception, outperforming both one-step and multi-step methods. \textbf{Zoom in for details.}}
    \label{fig: teaser}
    \vspace{30pt}
\end{strip}

\renewcommand{\thefootnote}{}
\footnotetext{\dag \ Corresponding author.}
\footnote{This work was completed during an internship at vivo.}

\begin{abstract}
Owing to the robust priors of diffusion models, recent approaches have shown promise in addressing real-world super-resolution (Real-SR). However, achieving semantic consistency and perceptual naturalness to meet human perception demands remains difficult, especially under conditions of heavy degradation and varied input complexities. To tackle this, we propose Hero-SR, a one-step diffusion-based SR framework explicitly designed with human perception priors. Hero-SR consists of two novel modules: the Dynamic Time-Step Module (DTSM), which adaptively selects optimal diffusion steps for flexibly meeting human perceptual standards, and the Open-World Multi-modality Supervision (OWMS), which integrates guidance from both image and text domains through CLIP to improve semantic consistency and perceptual naturalness. Through these modules, Hero-SR generates high-resolution images that not only preserve intricate details but also reflect human perceptual preferences. Extensive experiments validate that Hero-SR achieves state-of-the-art performance in Real-SR.
The code will be publicly available upon paper acceptance.
\end{abstract}

\section{Introduction}
Image super-resolution (SR) reconstructs high-resolution (HR) images from low-resolution (LR) inputs and is critical in fields such as computational photography, video surveillance, and media entertainment, where perceptually accurate visuals are essential~\cite{photograph, chen2019camera, video}.
In these applications, perceptual quality directly affects user interpretation and interaction with the content, impacting usability and user experience. 
However, achieving SR that aligns with high perceptual quality remains a challenge, particularly in real-world SR (Real-SR) tasks with complex degradations like noise and compression~\cite{RealESRGAN}.

Traditional pixel-based methods minimize pixel-level distortions but often result in overly smooth images~\cite{SRCNN, dong2016accelerating, gu2019blind}. GAN-based approaches enhance realism but introduce unnatural artifacts~\cite{RealESRGAN, wang2020deep, ESRGAN, LDL}.
Recently, diffusion-based~\cite{DDPM} SR methods have gained attention for their strong priors. Approaches such as StableSR~\cite{StableSR}, DiffBIR~\cite{diffbir}, and SeeSR~\cite{seesr} use pre-trained diffusion models along with guidance mechanisms such as ControlNet~\cite{controlnet} to improve SR quality. As diffusion models often require hundreds of iterative steps, methods such as ADDSR~\cite{addsr}, OSEDiff~\cite{osediff}, and S3Diff~\cite{s3diff} apply distillation techniques to reduce the computational cost by using the LR image as a starting point~\cite{img2img-turbo} and specialized losses to minimize the number of steps.
Despite these improvements, 
these methods struggle to meet human perception demands for better photo-realistic image super-resolution effects.

In this paper, we interpret the concept of human perception~\cite{lpips} in SR from two core factors
: semantic consistency and perceptual naturalness. 
Semantic consistency ensures that generated images maintain meaningful content. Methods like SeeSR~\cite{seesr}, PASD~\cite{pasd}, and SUPIR~\cite{supir} apply various forms of semantic guidance, such as tags, high-level semantic cues, and multimodal textual descriptions. However, these methods often lack the explicit semantic supervision essential for diffusion models to align effectively with semantic consistency.
Perceptual naturalness, on the other hand, requires that generated images not only follow general distribution but also align with human perceptual standards. Studies in image quality assessment, such as CLIPIQA~\cite{clipiqa} and Q-Align~\cite{q-align}, have shown that simply approximating statistical distributions is insufficient; Human-centered evaluations are crucial to align image quality with perceptual standards. 
However, current SR methods often overlook semantic consistency and perceptual naturalness, leading to images that fall short of human perception standards for coherence and realism.

To address these issues, we propose \textbf{Hero-SR}, a one-step diffusion-based super-resolution framework with \textbf{H}uman-p\textbf{er}ception pri\textbf{o}rs, specifically designed to improve semantic consistency and perceptual naturalness. Hero-SR consists of two novel modules: the Dynamic Time-Step Module (DTSM) and Open-World Multi-modality Supervision (OWMS).
First, the DTSM dynamically selects the optimal time-step based on image-specific features, precisely restoring intricate details. Unlike previous methods~\cite{addsr, osediff, s3diff} that use a fixed starting point from pure noise, DTSM adaptively chooses a starting step from a flexible range by analyzing image degradation and structural complexity. Leveraging a feature extraction network and the Gumbel-Softmax method, DTSM aligns the denoising process with visual details, flexibly meeting human perceptual standards.
Second, OWMS improves semantic consistency and perceptual naturalness by integrating CLIP multimodal guidance~\cite{clip}, aligning SR outputs with both text and image information. In the text domain, perceptual attribute prompts (e.g., quality, sharpness, clarity) guide the model toward criteria that reflect human preferences. In the image domain, the image encoder of CLIP~\cite{clip} extracts contextual features, enforcing semantic consistency across generated outputs.

Hero-SR integrates DTSM and OWMS to apply human perception priors throughout the SR process, addressing key aspects such as semantic consistency and perceptual naturalness. As shown in Figure~\ref{fig: teaser}, extensive experiments demonstrate the effectiveness and flexibility of Hero-SR. 
The contributions of our work can be summarized as follows:

\begin{itemize}
    \item We introduce Hero-SR, a one-step diffusion-based super-resolution framework with human perception priors. To the best of our knowledge, we are the first to incorporate multimodal models into the training of Real-SR tasks.
    \item Hero-SR integrates two novel modules, DTSM and OWMS, to enforce semantic consistency and perceptual naturalness throughout the SR process, ensuring perceptually accurate restorations.
    \item Hero-SR achieves state-of-the-art performance, outperforming existing one-step and multi-step methods in both quantitative and qualitative evaluations.
\end{itemize}

\begin{figure*}[t]
\centering
\includegraphics[width=1.0\linewidth]{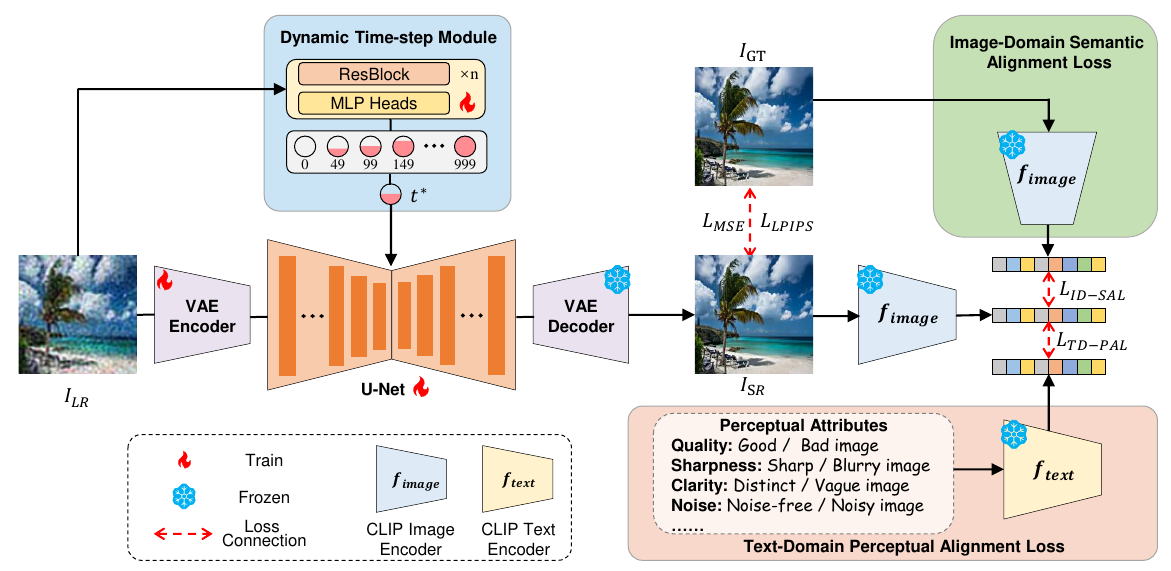}
\caption{Training framework of Hero-SR. Hero-SR incorporates a Dynamic Time-step Module to adaptively determine the optimal time-step \( t^* \) based on the input image \( I_{LR} \), flexibly meeting human perceptual standards. Both \( I_{LR} \) and \( t^* \) are then input to the diffusion network to generate the restored image \( I_{SR} \). Text-Domain Perceptual Alignment Loss and Image-Domain Semantic Alignment Loss ensure semantic consistency and perceptual naturalness, aligning outputs with human perception.}
\label{fig: framework}
\end{figure*}

\section{Related Work}
\subsection{Real-world Image Super-Resolution}
Deep learning has driven advances in SR, beginning with methods like SRCNN~\cite{SRCNN}, which introduced deep neural networks for SR. Subsequent architectures, such as ResNet and Transformer models~\cite{Swinir, gu2019blind, wang2020deep, dong2016accelerating}, emphasize fidelity through pixel-level losses. However, these methods often yield overly smooth images that lack the details essential for human perception alignment.
Such artifacts can negatively impact the practical usability of SR models, especially in applications like media and surveillance, where fine details are crucial.
These limitations are amplified in real-world super-resolution (Real-SR) tasks~\cite{RealSR}, where images are degraded by noise, compression, and other distortions. The challenge in Real-SR is to restore fine details while ensuring semantic consistency and perceptual naturalness, requiring models that can handle complex degradations and maintain visual fidelity.
GAN-based methods~\cite{GAN, Bsrgan, RealESRGAN} incorporate adversarial training to generate finer details. While effective in enhancing realism, GANs frequently introduce unnatural artifacts due to training instability, disrupting semantic consistency~\cite{LDL}. Additionally, GAN-based SR models struggle to preserve coherent global structures, limiting their ability to meet human perception standards~\cite{RealESRGAN}. These limitations underscore the need for generative models with stronger priors. Recently, diffusion models have shown strong potential in generating high-quality images with enhanced detail and coherence.
\subsection{Diffusion-based Real-SR}
Diffusion models employ Markov processes to generate complex data distributions, with foundational models like DDPM~\cite{DDPM} and DDIM~\cite{ddim} establishing the groundwork. The Latent Diffusion Model~\cite{ldm} further improves computational efficiency, enabling large-scale pretrained models such as Stable Diffusion~\cite{sdxl}. 
Extensions like ControlNet~\cite{controlnet} provide added control over the generation process, enhancing applications of diffusion models in restoration and editing.
In SR tasks, diffusion-based methods generally fall into three categories. The first approach~\cite{chung2022come,lugmayr2022repaint, sdedit, avrahami2022blended} modifies pretrained diffusion models with gradient descent but is constrained by reliance on predefined degradation models, limiting adaptability in real-world scenarios. The second approach, including methods like ResShift~\cite{resshift} and SinSR~\cite{sinsr}, trains models from scratch on paired data, but results are limited by data diversity and scale. 
Consequently, the adaptability of these models to complex, real-world degradation patterns remains limited, as they often struggle to adapt to challenging conditions.
The third and most common approach leverages pretrained diffusion models with ControlNet~\cite{controlnet} to generate high-quality SR outputs from LR inputs. Models like StableSR~\cite{StableSR}, SeeSR~\cite{seesr}, DiffBIR~\cite{diffbir}, and others~\cite{pasd,supir} improve upon this approach by incorporating architectural and semantic guidance, yielding visually enhanced outputs. Diffusion-based SR methods typically require numerous sampling steps, reducing practical efficiency.
To address this limitation, recent diffusion-based SR methods like ADDSR~\cite{addsr}, S3Diff~\cite{s3diff}, and OSEDiff~\cite{osediff} incorporate adversarial distillation and score-matching to accelerate inference. However, diffusion-based SR methods still fall short of fully meeting human perception standards, especially in semantic consistency, and perceptual naturalness. This highlights the need for SR methods better aligned with human visual expectations.

\section{Methodology}

\subsection{Framework Overview}

Hero-SR is a one-step diffusion-based super-resolution framework with human perception priors, with two core modules: the Dynamic Time-Step Module (DTSM) to flexibly meet human perceptual standards and the Open-World Multi-modality Supervision (OWMS) for perceptual and semantic alignment. Hero-SR is built on the Stable Diffusion model~\cite{ldm}, comprising a VAE encoder $\mathcal{E}$, a U-Net $\mathcal{U}$, and a VAE decoder $\mathcal{D}$.

As shown in Fighure~\ref{fig: framework}, given a low-resolution input $I_{\text{LR}}$, DTSM adaptively selects an optimal time-step $t^* = \text{DTSM}(I_{\text{LR}})$. The VAE encoder encodes $I_{\text{LR}}$ into a latent representation $z_{\text{LR}} = \mathcal{E}(I_{\text{LR}})$, which is then processed by the U-Net at $t^*$ to produce an enhanced latent representation $z_{\text{SR}} = \mathcal{U}(z_{\text{LR}}, t^*)$. Finally, the VAE decoder reconstructs the high-resolution output $I_{\text{SR}} = \mathcal{D}(z_{\text{SR}})$. Low-rank adaptation (LoRA)~\cite{LoRA} is applied, with the VAE decoder frozen during training to maintain its generative capacity.

\begin{figure}[t]
\centering
\includegraphics[width=1.0\linewidth]{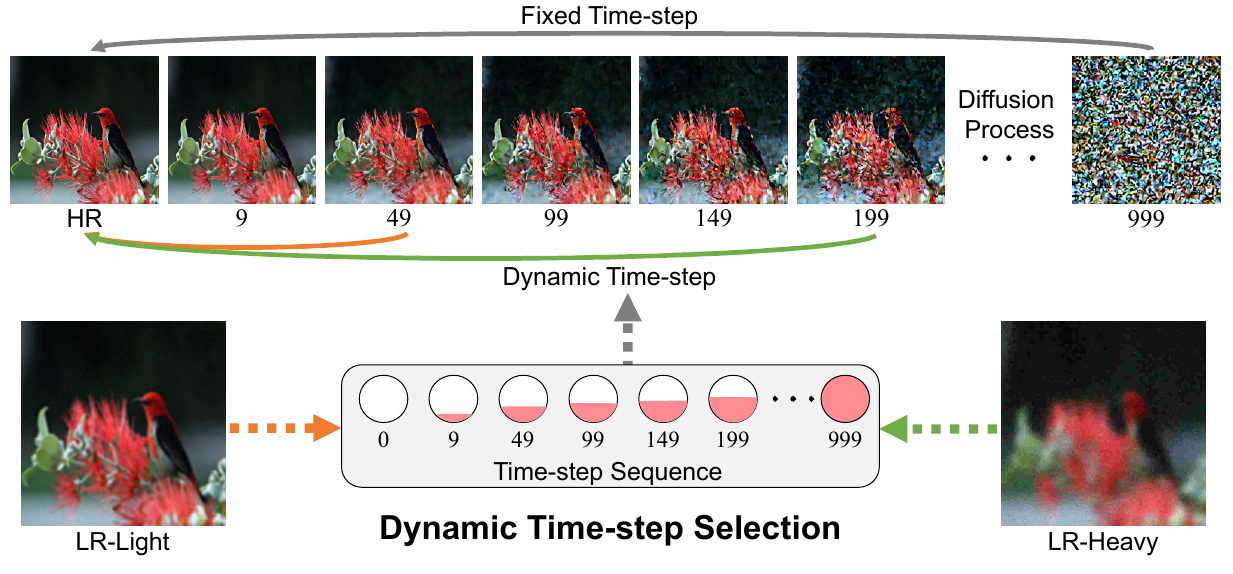}
\caption{The time-step selection process of DTSM. Previous one-step methods use a fixed starting time-step from pure noise, while DTSM adaptively selects a dynamic starting time-step based on the input image to better align with the diffusion process.}
\label{fig:timestep}
\end{figure}

\subsection{Dynamic Time-Step Module}

In Real-SR tasks, the degradation of input images varies widely, leading to a range of structural complexities~\cite{RealSR}. As shown in Figure~\ref{fig:timestep}, previous one-step diffusion-base SR methods~\cite{addsr,s3diff, osediff} with a fixed starting time-step, such as step 999, fail to account for this variation, limiting the restoration of rich details. The Dynamic Time-Step Module (DTSM) addresses this by adaptively selecting the optimal time-step based on the degradation level and complexity of the input image, thus improving detail restoration to flexibly meet perceptual standards.

Diffusion models operate through progressive denoising~\cite{DDPM}, with each time-step corresponding to a noise level~\cite{luo2023image}. To allow DTSM to adapt to different levels of input complexity, we select a time-step candidate subset \( S \) from the original diffusion sequence: 
\[
S \subseteq \{x \in \mathbb{Z} \mid 0 \leq x \leq 999\},
\]
where each element in \( S \) represents a specific noise level across the diffusion trajectory. 

For a given low-resolution input \( I_{\text{LR}} \), DTSM extracts features relevant to degradation and complexities to guide time-step selection. First, \( I_{\text{LR}} \) is processed through convolutional layers to capture localized feature patterns:
\begin{equation} 
f_{shallow} = \textnormal{Conv}(I_{\text{LR}}).
\end{equation} 
These features are further refined through a series of residual blocks~\cite{resnet} to model more complex characteristics:
\begin{equation} 
f_{deep}  = \textnormal{ResBlocks}_{n}(f_{shallow} ),
\end{equation} 
where \( n \) is the number of residual blocks. This output \( f_{deep}  \) is then flattened and passed through a multi-layer perceptron (MLP), yielding a compact feature vector \( v \):
\begin{equation} 
v = \textnormal{MLP}(\text{Flatten}(f_{deep}  )).
\end{equation} 

To select the optimal time-step \( t^* \), we use the Gumbel-Softmax trick~\cite{gumbel-softmax}, which enables differentiable selection during training. The time-step \( t^* \) is computed as:
\begin{equation} 
t^* = \textnormal{Gumbel-Softmax}(v, S).
\end{equation} 

By aligning the denoising process with both structural complexity and degradation characteristics of the input, DTSM effectively balances detail restoration with perceptual naturalness, flexibly aligning with human perception requirements.

\subsection{Open-World Multi-modality Supervision}

To address the limitations of traditional loss functions and better align with human perception, we propose an Open-World Multi-modality Supervision strategy (OWMS). This approach leverages the powerful multimodal capabilities of CLIP~\cite{clip}, a model pre-trained on large-scale datasets to establish strong visual-textual associations, achieving an open-world level of perceptual understanding~\cite{clipiqa}.
This shared space enables effective alignment through two main components: the Text-Domain Perceptual Alignment Loss (TD-PAL), which guides perceptual alignment, and the Image-Domain Semantic Alignment Loss (ID-SAL), which enforces semantic consistency.

\begin{table}[t]
\caption{Perceptual attributes and their corresponding prompts. We select key perceptual attributes closely aligned with human perception and apply respective positive and negative prompts.}
\resizebox{\linewidth}{!}{
\begin{tabular}{c|c|c}
\toprule
Perceptual Attributes & Positive Prompts & Negative Prompts \\
\midrule
Quality & Good image & Bad image \\
Sharpness & Sharp image & Blurry image \\
Edge Clarity & Sharp edges & Blurry edges \\
Resolution & High resolution image & Low resolution image \\
Noise & Noise-free image & Noisy image \\
Clarity & Distinct image & Vague image \\
\bottomrule
\end{tabular}}

\label{tab:perceptual_attribution}
\end{table}

\subsubsection{Text-Domain Perceptual Alignment Loss}

Text-Domain Perceptual Alignment Loss (TD-PAL) aligns restored images with human-perceptual standards by focusing on \( n \) perceptual attributes, each represented by a positive and negative prompt pair~\cite{clipiqa}, as shown in Table~\ref{tab:perceptual_attribution}.
By adjusting these attributes, TD-PAL enhances the perceptual quality of restored images, aligning them more closely with human expectations.

For a restored image \( I_{\text{SR}} \), we compute its embedding \( \mathbf{e}_{\text{SR}} \) using image encoder of CLIP \( f_{\text{image}} \):
\begin{equation} 
\mathbf{e}_{\text{SR}} = f_{\text{image}}(I_{\text{SR}}).
\label{eq:e_sr}
\end{equation} 

Subsequently, we encode predefined prompts using the text encoder. 
Each attribute has a positive prompt \( T_i^p \) and a negative prompt \( T_i^n \). Using  text encoder of CLIP \( f_{\text{text}} \), we obtain their embeddings:
\begin{equation} 
\mathbf{e}_{\text{text}}^{(i,p)} = f_{\text{text}}(T_i^p), \quad \mathbf{e}_{\text{text}}^{(i,n)} = f_{\text{text}}(T_i^n),
\end{equation} 
where \( \mathbf{e}_{\text{text}}^{(i,p)} \) and \( \mathbf{e}_{\text{text}}^{(i,n)} \) denote the embeddings of the positive and negative prompts for the \( i \)-th attribute, respectively.

To assess alignment in the perceptual attributes, we compute the cosine similarity between the image embedding $\mathbf{e}_{\text{SR}} $ and each text embedding,    \( \mathbf{e}_{\text{text}}^{(i,p)} \) and \( \mathbf{e}_{\text{text}}^{(i,n)} \):
\begin{equation} 
s_i^{(p)} = \frac{\mathbf{e}_{\text{SR}} \odot \mathbf{e}_{\text{text}}^{(i,p)}}{\|\mathbf{e}_{\text{SR}}\| \cdot \|\mathbf{e}_{\text{text}}^{(i,p)}\|}, \quad s_i^{(n)} = \frac{\mathbf{e}_{\text{SR}} \odot \mathbf{e}_{\text{text}}^{(i,n)}}{\|\mathbf{e}_{\text{SR}}\| \cdot \|\mathbf{e}_{\text{text}}^{(i,n)}\|},
\end{equation} 
where \( s_i^{(p)} \) and \( s_i^{(n)} \) represent the cosine similarities between the image embedding \( \mathbf{e}_{\text{SR}} \) and each positive and negative prompt for the \( i \)-th attribute, respectively.

We apply softmax normalization for stability:
\begin{equation} 
\hat{s}_i^{(p)} = \frac{e^{s_i^{(p)}}}{e^{s_i^{(p)}} + e^{s_i^{(n)}}},
\end{equation} 
where \( \hat{s}_i^{(p)} \) represents the normalized similarity score for the positive prompt of the \( i \)-th attribute, reflecting the alignment of \( I_{\text{SR}} \) with perceptual attributes and enabling stable comparisons between positive and negative prompts for consistent alignment across attributes.

TD-PAL is then defined as:
\begin{equation} 
\mathcal{L}_{\text{TD-SAL}} = 1 - \frac{1}{n} \sum_{i=1}^{n} \hat{s}_i^{(p)},
\end{equation} 
encouraging the alignment of \( I_{\text{SR}} \) with human-perceptual standards across each attribute. This alignment enhances the perceptual quality of the generated images, making them more attuned to human quality assessments.

\subsubsection{Image-Domain Semantic Alignment Loss }
Diffusion models differ from traditional SR approaches by relying on semantic information to guide image generation. However, existing methods~\cite{seesr, pasd, supir} focus on semantic guidance, neglecting the importance of semantic supervision. To address this gap, we propose the Image-Domain Semantic Alignment Loss (ID-SAL) to improve the generative ability of the model through semantic-level alignment.

ID-SAL enforces semantic consistency by aligning restored images with ground truth (GT) images. For a restored image \( I_{\text{SR}} \) and its GT image \( I_{\text{GT}} \), we use the CLIP image encoder to compute their embeddings in the semantic space. Since \( \mathbf{e}_{\text{SR}} \), the embedding for \( I_{\text{SR}} \), has already been computed in Equation~\eqref{eq:e_sr}, we compute only the embedding for \( I_{\text{GT}} \), denoted as \( \mathbf{e}_{\text{GT}} \), as follows:
\begin{equation} 
\mathbf{e}_{\text{GT}} = f_{\text{image}}(I_{\text{GT}}).
\end{equation} 

Next, we calculate the cosine similarity between $\mathbf{e}_{\text{SR}}$  and $\mathbf{e}_{\text{GT}}$ to assess sematic alignment:
\begin{equation} 
s = \frac{\mathbf{e}_{\text{SR}} \odot \mathbf{e}_{\text{GT}}}{\|\mathbf{e}_{\text{SR}}\| \cdot \|\mathbf{e}_{\text{GT}}\|},
\end{equation} 
where \( s \in [-1, 1] \) denotes the semantic alignment score, with values closer to 1 indicating higher alignment in semantic space. This score quantifies how well the restored image \( I_{\text{SR}} \) preserves the semantic content of its ground-truth counterpart \( I_{\text{GT}} \).

ID-SAL is then defined as:
\begin{equation} 
\mathcal{L}_{\text{ID-SAL}} = 1 - s,
\end{equation} 
which drives the restored image to maintain semantic fidelity with its GT counterpart. This alignment improves the ability of the model to produce semantically consistent outputs, enhancing both perceptual coherence and fidelity for diverse real-world SR inputs.

\subsection{Total Loss Function}
The total loss combines multiple objectives to balance fidelity, perceptual alignment, and semantic consistency:
\begin{equation}
\mathcal{L}_{\text{total}} = \lambda_1 \mathcal{L}_{\text{MSE}} + \lambda_2 \mathcal{L}_{\text{LPIPS}} + \lambda_3 \mathcal{L}_{\text{TD-PAL}} + \lambda_4 \mathcal{L}_{\text{ID-SAL}},
\end{equation}
where, \(\lambda_{i}\) corresponds to \(\mathcal{L}_{i}\) for \(i = 1, 2, 3, 4\), representing \(\mathcal{L}_{\text{MSE}}\), \(\mathcal{L}_{\text{LPIPS}}\), \(\mathcal{L}_{\text{TD-PAL}}\), and \(\mathcal{L}_{\text{ID-SAL}}\) respectively. 
This combination ensures that Hero-SR meets human perception criteria, achieving high-quality detail restoration, semantic consistency, and perceptual naturalness in SR tasks. 

\begin{table*}[t]
\centering
\small
\caption{Quantitative comparison with \textbf{one-step} diffusion methods on both synthetic and real-world benchmarks. The best and second best results of each metric are highlighted in {\color[HTML]{FF0000}\textbf{red}} and {\color[HTML]{0000FF}\textbf{blue}}, respectively.}
\resizebox{\linewidth}{!}{
\begin{tabular}{c|c|ccccccccc}
\toprule
Datasets & Methods & PSNR$\uparrow$ & SSIM$\uparrow$ & LPIPS$\downarrow$ & MUSIQ$\uparrow$ & HyperIQA$\uparrow$ & TOPIQ$\uparrow$ & TRES$\uparrow$ & ARNIQA$\uparrow$ & Q-Align$\uparrow$ \\
\midrule

 \multirow{5}{*}{DIV2K}& ADDSR & 23.2604 & 0.5902 & 0.3623 & 63.3961 & 0.5829 & 0.5730 & 73.3587 & 0.7107 & 3.2480 \\
 & S3Diff & 23.5164 & 0.5949 & {\color[HTML]{FF0000} \textbf{0.2581}} & 68.0107 & 0.6376 & 0.6342 & 80.7641 & {\color[HTML]{0000FF} \textbf{0.7209}} & 3.7949 \\
 
 & OSEDiff & 23.3820 & 0.6009 & 0.3173 & {\color[HTML]{0000FF} \textbf{69.1473}} & {\color[HTML]{0000FF} \textbf{0.6421}} & {\color[HTML]{0000FF} \textbf{0.6459}} & {\color[HTML]{0000FF} 
 \textbf{82.4990}} & {\color[HTML]{0000FF} \textbf{0.7209}} & {\color[HTML]{FF0000} \textbf{4.0781}} \\
 & SinSR & {\color[HTML]{FF0000} \textbf{24.4111}} & {\color[HTML]{0000FF} \textbf{0.6017}} & 0.3239 & 62.7990 & 0.5797 & 0.5721 & 72.5458 & 0.6659 & 3.1895 \\
 
 & Hero-SR (Ours)& {\color[HTML]{0000FF} \textbf{24.3663}} & {\color[HTML]{FF0000} \textbf{0.6257}} & {\color[HTML]{0000FF} \textbf{0.3111}} & {\color[HTML]{FF0000} \textbf{69.8524}} & {\color[HTML]{FF0000} \textbf{0.6711}} & {\color[HTML]{FF0000} \textbf{0.6948}} & {\color[HTML]{FF0000} \textbf{87.3938}} & {\color[HTML]{FF0000} \textbf{0.7255}} & {\color[HTML]{0000FF} \textbf{3.9968}} \\
\midrule

 \multirow{5}{*}{DrealSR} & ADDSR & 27.7707 & 0.7722 & 0.3196 & 60.8542 & 0.5797 & 0.5688 & 71.7246 & 0.6654 & 3.2578 \\
 & S3Diff & 27.3852 & 0.7468 & {\color[HTML]{0000FF} \textbf{0.3130}} & 64.1622 & 0.6053 & 0.6053 & 75.6122 & 0.6784 & 3.6094 \\
 & OSEDiff & 27.6269 & {\color[HTML]{0000FF} \textbf{0.7740}} & 0.3159 & {\color[HTML]{0000FF} \textbf{66.3766}} & {\color[HTML]{0000FF} \textbf{0.6287}} &{\color[HTML]{0000FF}{\textbf{0.6220}}} & {\color[HTML]{0000FF} \textbf{79.4500}} & {\color[HTML]{0000FF} \textbf{0.6833}} & {\color[HTML]{FF0000} \textbf{3.6855}} \\
 & SinSR & {\color[HTML]{0000FF} \textbf{28.3578}} & 0.7518 & 0.3659 & 55.6310 & 0.5182 & 0.5193 & 61.4332 & 0.5985 & 3.1191 \\
 
& Hero-SR (Ours)& {\color[HTML]{FF0000} \textbf{28.8962}} & {\color[HTML]{FF0000} \textbf{0.8016}} & {\color[HTML]{FF0000} \textbf{0.2933}} & {\color[HTML]{FF0000} \textbf{66.4874}} & {\color[HTML]{FF0000} \textbf{0.6434}} & {\color[HTML]{FF0000} \textbf{0.6622}} & {\color[HTML]{FF0000} \textbf{83.5888}} & {\color[HTML]{FF0000} \textbf{0.6913}} & {\color[HTML]{0000FF} \textbf{3.6302}} \\

\midrule
 \multirow{5}{*}{RealSR} & ADDSR & 24.7929 & 0.7077 & 0.3091 & 66.1849 & 0.6082 & 0.5991 & 79.9438 & 0.6923 & 3.4102 \\
 & S3Diff & 25.1930 & 0.7315 & {\color[HTML]{FF0000} \textbf{0.2707}} & 67.9144 & 0.6104 & 0.6137 & 78.7253 & 0.6969 & 3.6523 \\
 & OSEDiff & 24.8520 & 0.7218 & 0.3115 & {\color[HTML]{0000FF} \textbf{69.9864}} & {\color[HTML]{0000FF} \textbf{0.6469}} & {\color[HTML]{0000FF} \textbf{0.6506}} & {\color[HTML]{0000FF} \textbf{83.5311}} & {\color[HTML]{0000FF} \textbf{0.7013}} & {\color[HTML]{0000FF} \textbf{3.8047}} \\
 & SinSR & {\color[HTML]{FF0000} \textbf{26.3254}} & {\color[HTML]{0000FF} \textbf{0.7364}} & 0.3195 & 60.5987 & 0.5205 & 0.5184 & 67.8383 & 0.6435 & 3.1816 \\
& Hero-SR (Ours)& {\color[HTML]{0000FF} \textbf{25.8271}} & {\color[HTML]{FF0000} \textbf{0.7439}} & {\color[HTML]{0000FF} \textbf{0.2893}} & {\color[HTML]{FF0000} \textbf{70.0254}} & {\color[HTML]{FF0000} \textbf{0.6623}} & {\color[HTML]{FF0000} \textbf{0.6881}} & {\color[HTML]{FF0000} \textbf{88.5315}} & {\color[HTML]{FF0000} \textbf{0.7170}} & {\color[HTML]{FF0000} \textbf{3.8470}} \\

\bottomrule

\end{tabular}}

\label{table：one-step-result}
\end{table*}

\begin{table*}[t]
\centering
\small
\caption{Quantitative comparison with \textbf{multi-step} diffusion methods on both synthetic and real-world benchmarks. The best and second best results of each metric are highlighted in {\color[HTML]{FF0000}\textbf{red}} and {\color[HTML]{0000FF}\textbf{blue}}, respectively.}

\resizebox{\linewidth}{!}{
\begin{tabular}{c|c|c|ccccccccc}
\toprule
DataSet & Methods & Step & PSNR$\uparrow$ & SSIM$\uparrow$ & LPIPS$\downarrow$ & MUSIQ$\uparrow$ & HyperIQA$\uparrow$ & TOPIQ$\uparrow$ & TRES$\uparrow$ & ARNIQA$\uparrow$ & Q-Align$\uparrow$ \\
\midrule

\multirow{5}{*}{DIV2K}  
& StableSR & 200 &  23.2613 & 0.5726 & {\color[HTML]{0000FF} \textbf{0.3113}} & 65.9177 & 0.6130 & 0.5979 & 77.3719 & 0.6916 & 3.5273 \\
& DiffBIR & 50 & 23.4091 & 0.5732 & 0.3456 & 68.3954 & 0.6315 & 0.6344 & 80.9948 & 0.7002 & 3.7324 \\
& SeeSR & 50 & 23.6780 & 0.6043 & 0.3193 & {\color[HTML]{0000FF} \textbf{68.6721}} & {\color[HTML]{0000FF} \textbf{0.6679}} & {\color[HTML]{0000FF} \textbf{0.6857}} & {\color[HTML]{0000FF} \textbf{85.8015}} & {\color[HTML]{FF0000} \textbf{0.7302}} & {\color[HTML]{FF0000} \textbf{4.1211}} \\
& ResShift & 15 & {\color[HTML]{FF0000} \textbf{24.7538}} & {\color[HTML]{FF0000} \textbf{0.6300}} & 0.3649 & 60.0644 & 0.5564 & 0.5253 & 73.3462 & 0.6645 & 2.8613 \\
& Hero-SR (Ours)& 1 & {\color[HTML]{0000FF} \textbf{24.3663}} & {\color[HTML]{0000FF} \textbf{0.6257}} & {\color[HTML]{FF0000} \textbf{0.3111}} & {\color[HTML]{FF0000} \textbf{69.8524}} & {\color[HTML]{FF0000} \textbf{0.6711}} & {\color[HTML]{FF0000} \textbf{0.6948}} & {\color[HTML]{FF0000} \textbf{87.3938}} & {\color[HTML]{0000FF} \textbf{0.7255}} & {\color[HTML]{0000FF} \textbf{3.9968}} \\

\midrule
\multirow{5}{*}{DrealSR}  
& StableSR & 200 &28.0297 & 0.7536 & 0.3284 & 58.5118 & 0.5482 & 0.5323 & 66.7321 & 0.6254 & 3.0613 \\
& DiffBIR & 50 & 25.9304 & 0.6526 & 0.4518 & {\color[HTML]{0000FF} \textbf{65.6740}} & 0.6296 & 0.6149 & 77.8703 & 0.6546 & {\color[HTML]{0000FF} \textbf{3.5977}} \\
& SeeSR & 50 & 28.0719 & 0.7684 & 0.3174 & 65.0907 & {\color[HTML]{FF0000} \textbf{0.6642}} & {\color[HTML]{0000FF} \textbf{0.6574}} & {\color[HTML]{FF0000} \textbf{84.7264}} & {\color[HTML]{0000FF} \textbf{0.6883}} & 3.5879 \\
& ResShift & 15 & {\color[HTML]{0000FF} \textbf{28.8071}} & {\color[HTML]{FF0000} \textbf{0.8065}} & {\color[HTML]{0000FF} \textbf{ 0.3207}} & 53.3830 & 0.5031 & 0.4757 & 64.2898 & 0.6037 & 2.8145 \\
& Hero-SR (Ours)& 1 & {\color[HTML]{FF0000} \textbf{28.8962}} & {\color[HTML]{0000FF} \textbf{0.8016}} & {\color[HTML]{FF0000} \textbf{0.2933}} & {\color[HTML]{FF0000} \textbf{66.4874}} & {\color[HTML]{0000FF} \textbf{0.6434}} & {\color[HTML]{FF0000} \textbf{0.6622}} & {\color[HTML]{0000FF} \textbf{83.5888}} & {\color[HTML]{FF0000} \textbf{0.6913}} & {\color[HTML]{FF0000} \textbf{3.6302}} \\

\midrule
\multirow{5}{*}{RealSR}
& StableSR & 200 & 24.6451 & 0.7080 & 0.3002 & 65.8833 & 0.5796 & 0.5748 & 74.2591 & 0.6756 & 3.2764 \\
& DiffBIR & 50 & 24.2406 & 0.6649 & 0.3469 & 68.3388 & 0.6121 & 0.6052 & 78.9864 & 0.6717 & 3.6328 \\
& SeeSR & 50 & 25.1477 & 0.7210 & 0.3007 & {\color[HTML]{0000FF} \textbf{69.8191}} & {\color[HTML]{FF0000} \textbf{0.6748}} & {\color[HTML]{FF0000} \textbf{0.6891}} & {\color[HTML]{FF0000} \textbf{88.5903}} & {\color[HTML]{0000FF} \textbf{0.7155}} & {\color[HTML]{0000FF} \textbf{3.7148}} \\
& ResShift & 15 & {\color[HTML]{FF0000} \textbf{26.5344}} & {\color[HTML]{FF0000} \textbf{0.7636}} & {\color[HTML]{0000FF} \textbf{0.2964}} & 60.0152 & 0.5393 & 0.5215 & 73.7639 & 0.6647 & 3.0469 \\
& Hero-SR (Ours)& 1 & {\color[HTML]{0000FF} \textbf{25.8271}} & {\color[HTML]{0000FF} \textbf{0.7439}} & {\color[HTML]{FF0000} \textbf{0.2893}} & {\color[HTML]{FF0000} \textbf{70.0254}} & {\color[HTML]{0000FF} \textbf{0.6623}} & {\color[HTML]{0000FF} \textbf{0.6881}} & {\color[HTML]{0000FF} \textbf{88.5315}} & {\color[HTML]{FF0000} \textbf{0.7170}} & {\color[HTML]{FF0000} \textbf{3.8470}} \\

\bottomrule

\end{tabular}}

\label{table：multi-step-result}

\end{table*}

\section{Experiments}

\subsection{Experiments Setting}

\noindent \textbf{Training and Testing Datasets.}
We train the model on the LSDIR~\cite{lsdir} dataset, using the Real-ESRGAN~\cite{RealESRGAN} degradation pipeline to generate LR-HR training pairs. Testing is conducted on the StableSR~\cite{StableSR} test set, including synthetic and real data. The synthetic dataset consists of 3,000 images at 512$\times$512 resolution, with GT images randomly cropped from DIV2K-val~\cite{div2k} and degraded using Real-ESRGAN. Real data is sourced from RealSR~\cite{RealSR} and DRealSR~\cite{DrealSR}, containing 128$\times$128 and 512$\times$512 LR-HR pairs. This combination of synthetic and real-world test sets assesses the model on both controlled and unpredictable degradations, ensuring its robustness and generalization.

\noindent \textbf{Compared Methods.}
We compare our model with recent advanced diffusion model super-resolution methods, categorized into one-step (e.g., ADDSR~\cite{addsr}, S3Diff~\cite{s3diff}, OSEDiff~\cite{osediff}, SinSR~\cite{sinsr}) and multi-step approaches (e.g., StableSR~\cite{StableSR}, DiffBIR~\cite{diffbir}, SeeSR~\cite{seesr}, ResShit~\cite{resshift}). ResShift and its distilled one-step variant, SinSR, are trained from scratch, while other methods rely on pre-trained SD models. GAN-based methods such as SwinIR~\cite{Swinir}, BSRGAN~\cite{Bsrgan}, FeMaSR~\cite{femasr} and RealESRGAN~\cite{RealESRGAN} are presented in the Appendix for comparison.

\noindent \textbf{Evaluation Metrics.}
To comprehensively and accurately evaluate the performance of various methods, we employ a series of full-reference and no-reference metrics. PSNR and SSIM~\cite{ssim}, calculated on the Y channel in YCbCr space, serve as full-reference fidelity metrics, while LPIPS~\cite{lpips} is utilized as a full-reference perceptual quality metric. For no-reference image quality assessment, we employ advanced metrics such as MUSIQ~\cite{musiq}, HyperIQA~\cite{hyperiqa}, TOPIQ~\cite{topiq}, TRES~\cite{tres}, ARNIQA~\cite{arniqa}, and Q-Align~\cite{q-align}. These no-reference IQA methods are SOTA metrics, closely aligned with human subjective evaluations and perception. In particular, Q-Align, based on the LMM model, demonstrates exceptional evaluation capabilities.

\noindent \textbf{Implementation Details.}
Model training is conducted with the AdamW~\cite{adamw} optimizer at a learning rate of \(5 \times 10^{-5}\). Training is performed on 2 NVIDIA L40s GPUs for approximately 8 hours with a batch size of 2. SD-Turbo\footnote{https://huggingface.co/stabilityai/sd-turbo}~\cite{add} is used as a pre-trained diffusion model. The VAE encoder and U-Net network are fine-tuned using LoRA~\cite{LoRA} with a rank level of 16. The adaptive time-step module is trained from scratch with randomly initialized parameters. The weights of the losses $\lambda_1$, $\lambda_2$, $\lambda_3$, and $\lambda_4$ are set to 2, 5, 1, and 0.5, respectively. In TD-PAL and ID-SAL, the parameters of CLIP are frozen.

\begin{figure*}[t]
\centering
\resizebox{0.95\linewidth}{!}{\includegraphics{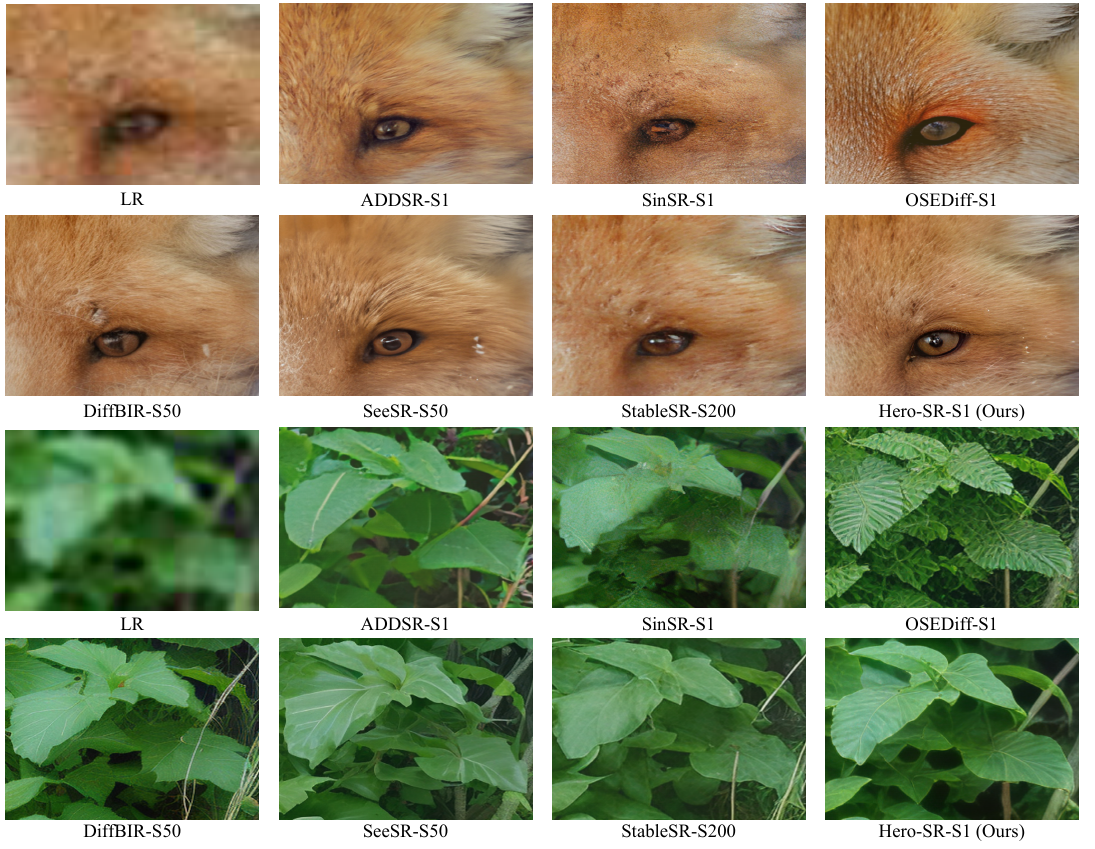}}
\caption{Qualitative comparison with one-step and multi-step methods. `S' indicates the number of diffusion steps. \textbf{Zoom in for details.}}
\label{fig:main_result}
\end{figure*}

\subsection{Comparison with State-of-the-Arts}

\subsubsection{Quantitative Comparisons.}
\noindent  \textbf{One-Step Methods.}
Table~\ref{table：one-step-result} presents the quantitative comparison between Hero-SR and other one-step methods. Key observations include: (1) Hero-SR consistently outperforms other methods across nearly all metrics, particularly on real-world datasets like DRealSR and RealSR. (2) Hero-SR achieves leading results in full-reference metrics, surpassing other methods in PSNR, SSIM, and LPIPS. SinSR attains a higher PSNR, likely due to its scratch-trained diffusion model, but underperforms on no-reference perceptual metrics. S3Diff shows a better LPIPS score but worse results on other no-reference metrics, likely due to its heavier LPIPS loss weighting during training.
(3) Hero-SR outperforms other methods across all datasets for no-reference perceptual metrics (e.g., MUSIQ, HyperIQA, TOPIQ, TRES, ARNIQA). For example, Hero-SR exceeds competitors by 7.0\% on the TOPIQ metric. These advanced no-reference metrics emphasize overall perceptual quality and align closely with human perception standards, highlighting the ability of Hero-SR to generate high-quality reconstructions that meet human visual expectations. 

\noindent  \textbf{Multi-Step Methods.}
Table~\ref{table：multi-step-result} provides the quantitative comparison between Hero-SR and multi-step methods, with key findings as follows: (1) As a one-step diffusion model, Hero-SR achieves competitive results with multi-step approaches across multiple datasets. (2) ResShift, which does not use a pre-trained diffusion model, shows relatively better performance on full-reference metrics like PSNR and SSIM but lower scores on no-reference metrics. However, compared to pretrained diffusion-based methods, Hero-SR achieves superior results on almost all full-reference fidelity metrics. (3) Hero-SR consistently ranks first or second across nearly all datasets in no-reference perceptual metrics, highlighting its strong alignment with human perceptual standards. These results demonstrate the ability of Hero-SR to capture visual qualities aligned with human judgment, such as naturalness and semantic consistency.

\subsubsection{Qualitative Comparisons.}
Figures~\ref{fig: teaser} and~\ref{fig:main_result} present visual comparison results. (1) In terms of texture restoration in the fox and owl case, Hero-SR generates more realistic details compared to other approaches. Compared to single-step methods, Hero-SR demonstrates clear advantages by producing more natural and perceptually aligned results. Compared to multi-step methods, Hero-SR produces more realistic texture details.
(2) In terms of semantic consistency in the leaf case, Hero-SR demonstrates superior semantic consistency by generating a coherent and complete leaf structure. Notably, Hero-SR not only preserves intricate details but also avoids introducing unnatural artifacts, thereby achieving a balance between local texture fidelity and global structural coherence. These results highlight the capabilities of Hero-SR across various scenarios, ranging from texture restoration to complex semantic alignment. \textbf{Additional visual results are provided in the appendix.}

\begin{table}[t]
\caption{Ablation study on the impact of different perceptual attributions.}
\centering
\resizebox{\linewidth}{!}{
\begin{tabular}{c|cccc}

\toprule
Perceptual Attributes & MUSIQ$\uparrow$  & HyperIQA$\uparrow$  & TOPIQ $\uparrow$ & Q-Align$\uparrow$  \\
\midrule \
w/o Quality & 65.4624 & 0.6316 & 0.6319 & 3.5633 \\
w/o Sharpness & 66.4843 & 0.6203 & 0.6589 & 3.5664 \\
w/o Edge Clarity & 65.6499 & 0.6360 & 0.6460 & 3.5534 \\
w/o Resolution & 66.3280 & 0.6315 & 0.6533 & 3.5936 \\
w/o Noise & 65.2068 & 0.6417 & 0.6524 & 3.5892 \\
w/o Clarity & 66.3015 & 0.6438 & 0.6579 & 3.6109 \\
All & \textbf{66.4874} & \textbf{0.6434} & \textbf{0.6622} & \textbf{3.6302} \\
\bottomrule

\end{tabular}}

\label{table:ablation_2}
\end{table}

\begin{table*}[ht]
\caption{Ablation study results on the effectiveness of the proposed DTSM, ID-SAL, and TD-PAL.}
\centering
\scriptsize
\resizebox{\linewidth}{!}{
\begin{tabular}{c|c|c|c|cccccc}
\toprule

Methods &DTSM & ID-SAL & TD-PAL &  MUSIQ$\uparrow$ & HyperIQA$\uparrow$ & TOPIQ$\uparrow$ & TRES$\uparrow$ & ARNIQA$\uparrow$ & Q-Align$\uparrow$ \\
\midrule

Variant-1 & $\times$ & $\checkmark$ & $\checkmark$& 66.4019 & 0.6361 & 0.6528 & 82.4065 & 0.6875 & 3.5282 \\
Variant-2 & $\checkmark$ & $\times$ & $\checkmark$ & 66.0649 & 0.6385 & 0.6617 & 83.4267 & 0.6609 & 3.5877 \\
Variant-3 & $\checkmark$ & $\checkmark$ & $\times$ & 60.7671 & 0.5849 & 0.5629 & 73.7752 & 0.6302 & 3.1945 \\
Hero-SR &$\checkmark$ & $\checkmark$ & $\checkmark$ & \textbf{66.4874} & \textbf{0.6434} & \textbf{0.6622} & \textbf{83.5888} & \textbf{0.6913} & \textbf{3.6302} \\
\bottomrule

\end{tabular}}

\label{table:ablation_1}
\end{table*}

\subsection{Ablation Study}

We first evaluate the effectiveness of the proposed DTSM and OWMS, with OWMS comprising the two components ID-SAL and TD-PAL, by testing Hero-SR with each module removed. Next, we analyze the impact of perceptual attributes within TD-PAL. Unless otherwise noted, all experiments are conducted on the DRealSR dataset with re-trained models, while holding all other settings constant.

\noindent \textbf{The Effectiveness of DTSM.} 
As shown in Table~\ref{table:ablation_1}, removing DTSM (Variant-1 vs. Hero-SR) results in a noticeable decline in no-reference perceptual metrics, including HyperIQA, TOPIQ, and Q-Align. These results underscore the critical function of DTSM in dynamically adjusting time-step to optimize perceptual quality. By adapting to image-specific features, DTSM effectively balances detail restoration with perceptual naturalness, aligning with human perceptual expectations.

\noindent \textbf{The Effectiveness of ID-SAL.}
As shown in Table~\ref{table:ablation_1}, the removal of ID-SAL (Variant-2 vs. Hero-SR) causes decreases in perceptual alignment metrics, highlighting the role of ID-SAL in maintaining semantic consistency. The reduction in Q-Align, a key metric for alignment with perceptual standards, emphasizes the contribution of ID-SAL to content coherence, ensuring generated images closely align with human perceptual expectations.

\noindent \textbf{The Effectiveness of TD-PAL.}
As shown in Table~\ref{table:ablation_1}, without TD-PAL (Variant-3 vs. Hero-SR), we observe noticeable declines in no-reference perceptual metrics, such as MUSIQ, TOPIQ, and TRES. These results suggest that TD-PAL is essential for enhancing perceptual naturalness, guiding the model to produce outputs that align well with perceptual standards.

\noindent \textbf{The Impact of Different Perceptual Attributions.}
The ablation study in Table~\ref{table:ablation_2} demonstrates the impact of individual perceptual attributes on the performance of Hero-SR, with some attributes contributing more than others. Excluding attributes like Quality and Noise led to marked declines in perceptual metrics; for example, removing Noise reduced HyperIQA, while omitting Quality notably lowered MUSIQ, underscoring the role of these attributes in achieving perceptual fidelity and naturalness. Including all perceptual attributes yields optimal performance, confirming that the combined use of all attributes is essential for aligning SR outputs with human perceptual standards and achieving high-quality, realistic results.

\section{Conclusion and Limitation}

We propose Hero-SR, a one-step diffusion-based super-resolution framework specifically designed with human perception priors to enhance semantic consistency and perceptual naturalness in real-world SR tasks. Hero-SR integrates two core modules: the Dynamic Time-Step Module (DTSM), which flexibly selects optimal diffusion steps to balance fidelity with perceptual standards, and the Open-World Multi-modality Supervision (OWMS), which leverages multimodal guidance from CLIP across image and text domains to reinforce semantic alignment with human visual preferences.
Through these modules, Hero-SR effectively captures fine details and produces high-resolution images closely aligned with human perceptual expectations. Extensive experiments demonstrate that Hero-SR achieves state-of-the-art performance across both real and synthetic datasets, surpassing existing one-step and multi-step methods in quantitative metrics and qualitative evaluation.

Hero-SR has certain limitations. Like other SD-based methods, it is constrained by the reconstruction capacity of the VAE, which restricts its ability to restore small structures, such as small-scale text and face. We aim to address these challenges in future work.

{
    \small
    \bibliographystyle{ieeenat_fullname}
    \bibliography{main}
}

\newpage
\appendix
\section*{Appendix}

\section{Comparison with GAN-based Methods}
We compare Hero-SR with four representative GAN-based Real-SR methods: BSRGAN~\cite{Bsrgan}, Real-ESRGAN~\cite{RealESRGAN}, SwinIR~\cite{Swinir}, and FeMaSR~\cite{femasr}, using three synthetic and real-world datasets~\cite{div2k,RealSR,DrealSR}. Quantitative and qualitative comparisons demonstrate that Hero-SR achieves superior perceptual consistency and generates more realistic textures, particularly in complex real-world scenarios.

\noindent \textbf{Quantitative Comparisons.} 
As shown in Table~\ref{tab:gan_result}, two key observations can be made: (1) GAN-based methods achieve higher fidelity metrics: GANs perform better on PSNR and SSIM. However, GAN-based methods are limited by their generative capacity and often fail to maintain high perceptual quality, falling short of aligning with human perception standards. 
(2) Hero-SR significantly outperforms GAN methods in perceptual quality: On no-reference perceptual metrics, such as MUSIQ~\cite{musiq} and Q-Align~\cite{q-align}, Hero-SR demonstrates a substantial advantage over all GAN-based methods. This improvement is attributed to the strong generative priors of diffusion models and the human perception design of Hero-SR, enabling exceptional perceptual alignment and naturalness.

\begin{figure*}[h]
\centering
\resizebox{1.0\linewidth}{!}{\includegraphics{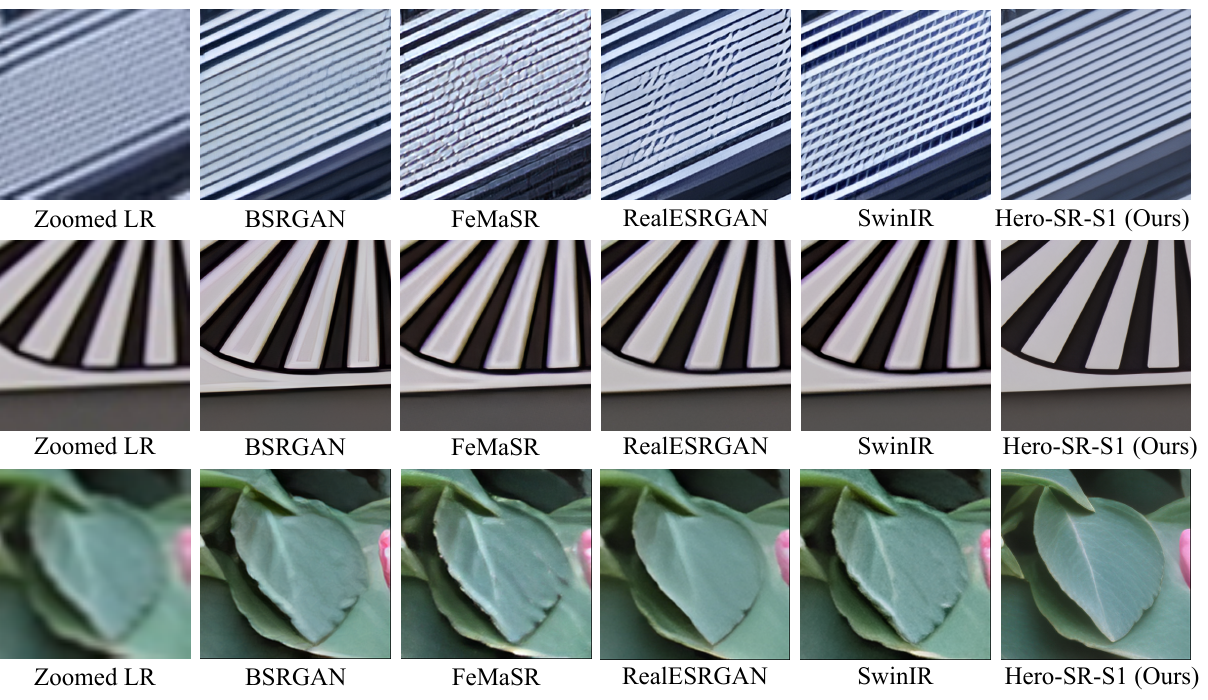}}
\caption{Qualitative comparison with GAN-base methods. \textbf{Zoom in for details.}}
\label{fig:gan_result}
\end{figure*}

\noindent \textbf{Qualitative Comparisons.} 
Figure~\ref{fig:gan_result} highlights the superiority of Hero-SR over GAN-based methods in texture restoration and semantic consistency. For instance, in the example of the blind, Hero-SR accurately restores high-frequency details and produces structured, natural textures. By contrast, while capturing some details, GAN methods fail to restore complex textures convincingly. In the leaf example, Hero-SR reconstructs a complete leaf structure with clearly defined vein patterns, achieving higher semantic consistency compared to GAN-based methods.

\section{Additional Visual Comparisons}
\begin{figure*}[ht]
\centering
\includegraphics[height=0.31\textheight]{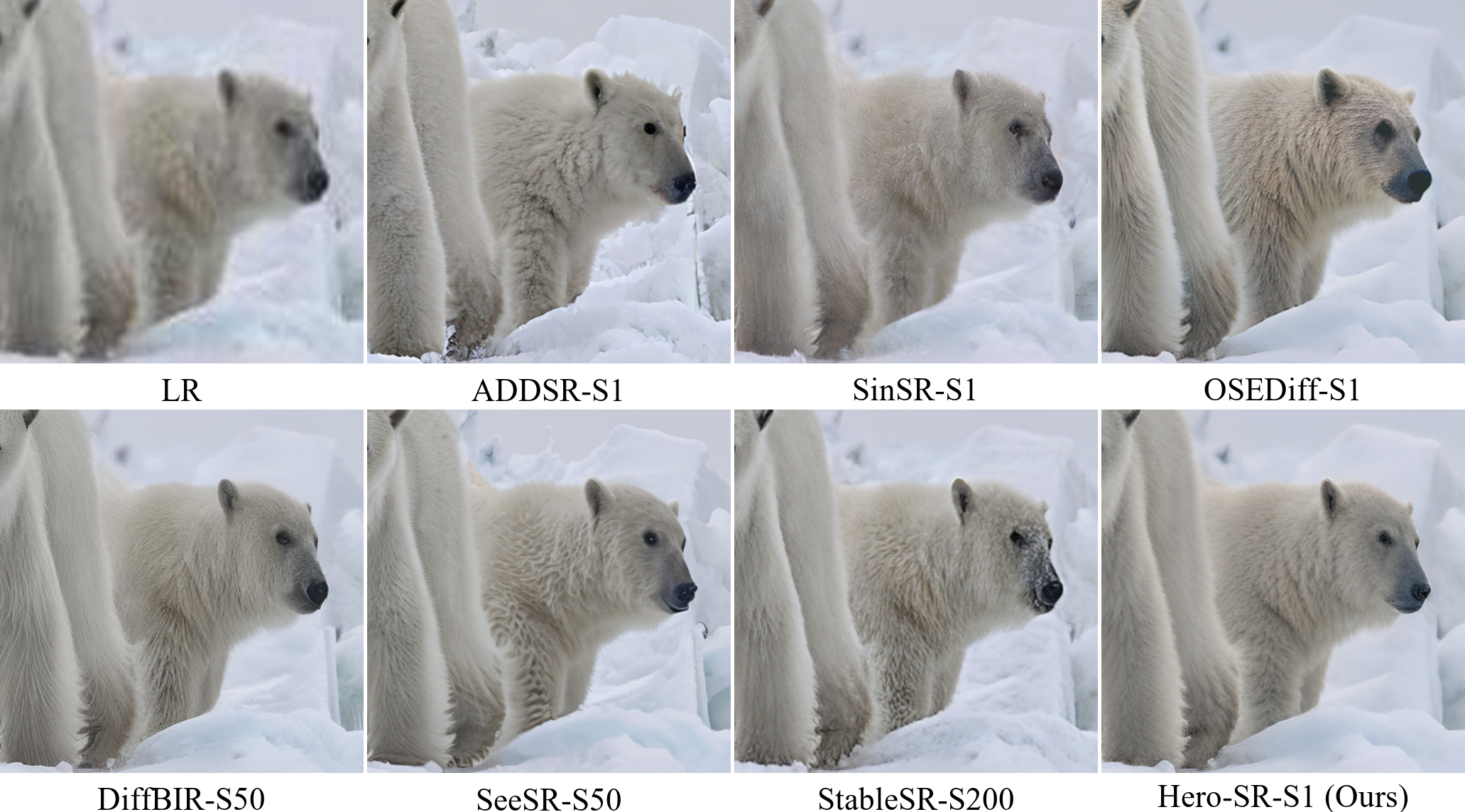}
\caption{Qualitative comparison with one-step and multi-step methods. `S' indicates the number of diffusion steps. \textbf{Zoom in for details.}}
\label{fig:result1}
\end{figure*}

\begin{figure*}[ht]
\centering
\includegraphics[height=0.31\textheight]{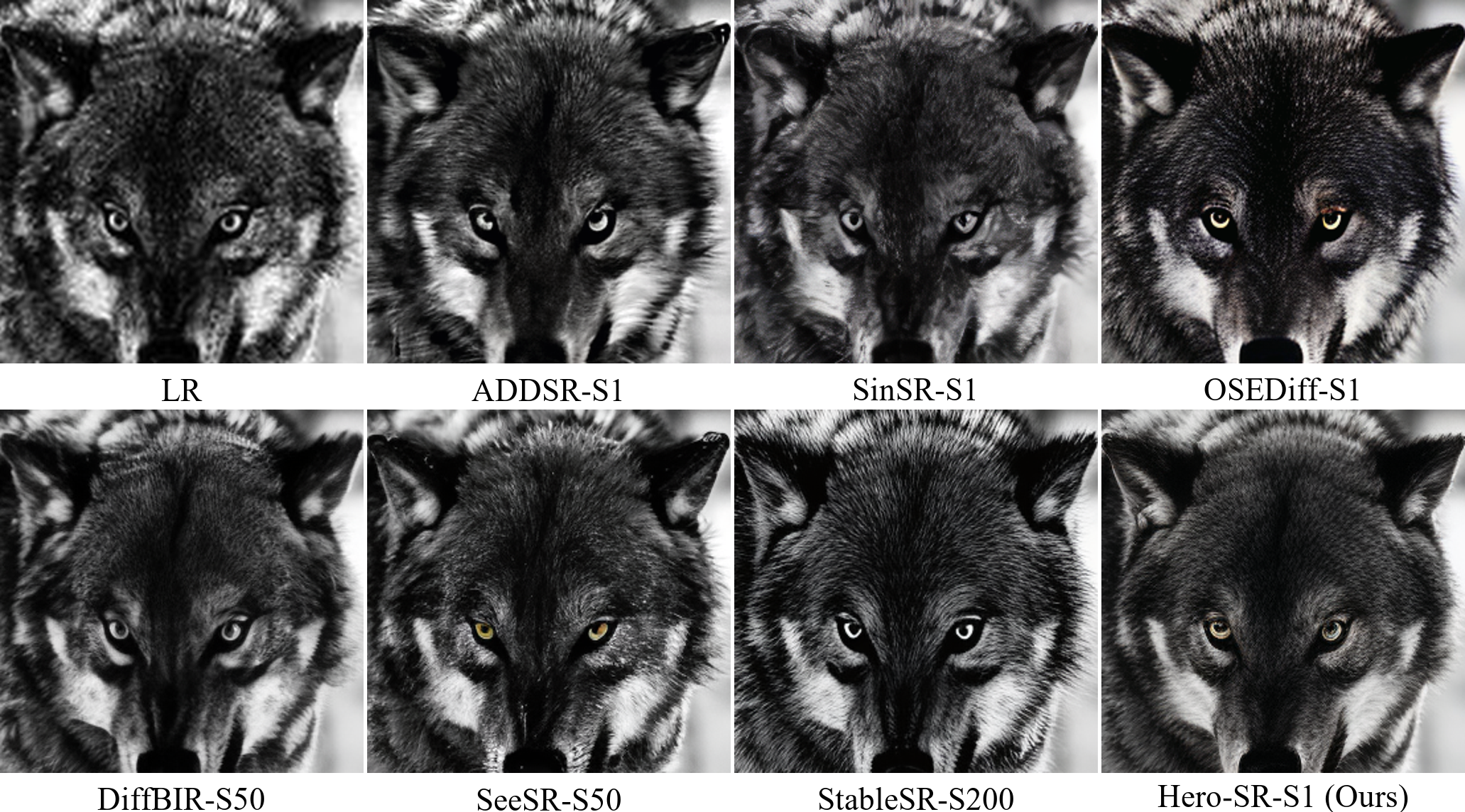}
\caption{Qualitative comparison with one-step and multi-step methods. `S' indicates the number of diffusion steps. \textbf{Zoom in for details.}}
\label{fig:result2}
\end{figure*}

\begin{figure*}[ht]
\centering
\includegraphics[height=0.31\textheight]{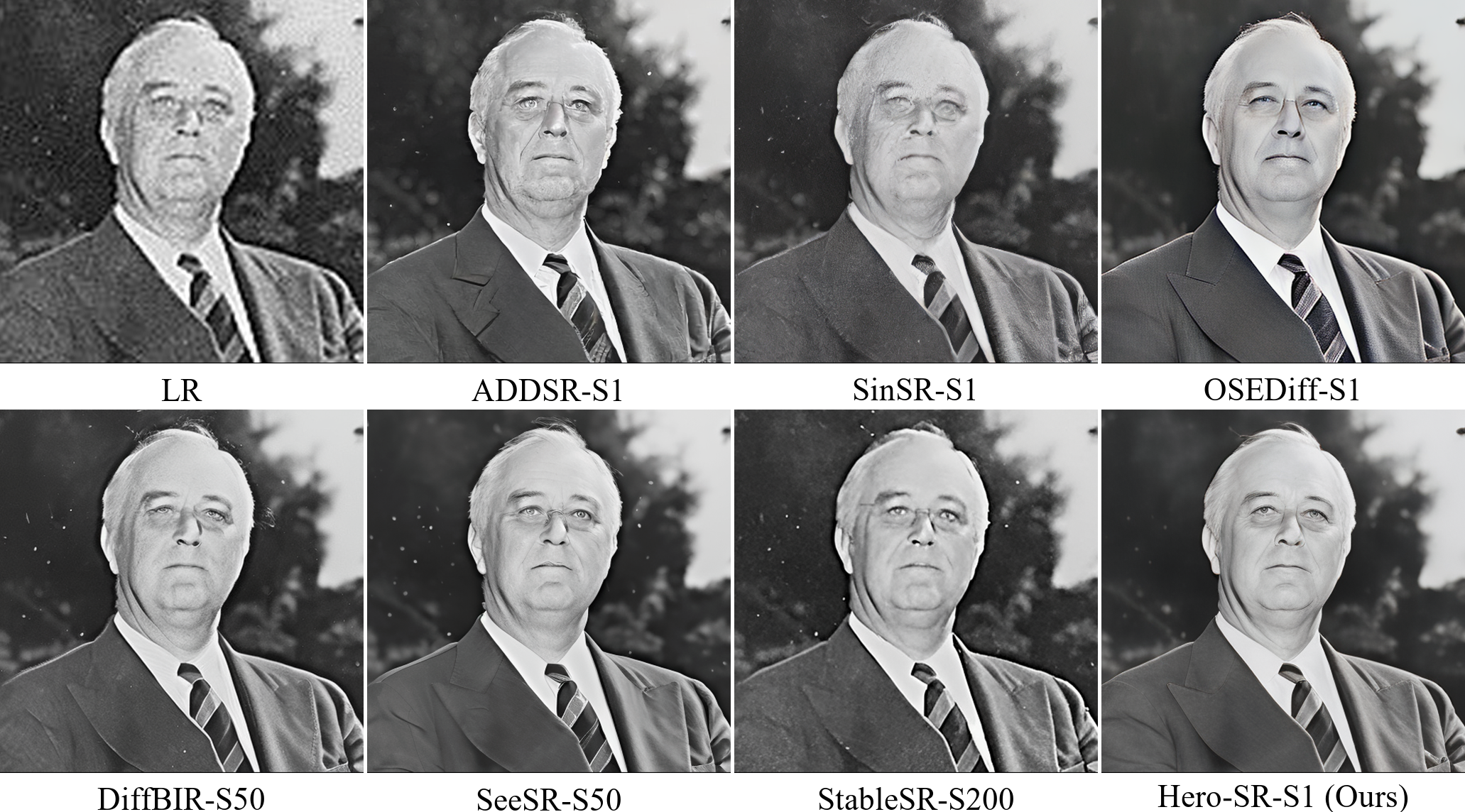}
\caption{Qualitative comparison with one-step and multi-step methods. `S' indicates the number of diffusion steps. \textbf{Zoom in for details.}}
\label{fig:result3}
\end{figure*}

\begin{figure*}[ht]
\centering
\includegraphics[height=0.31\textheight]{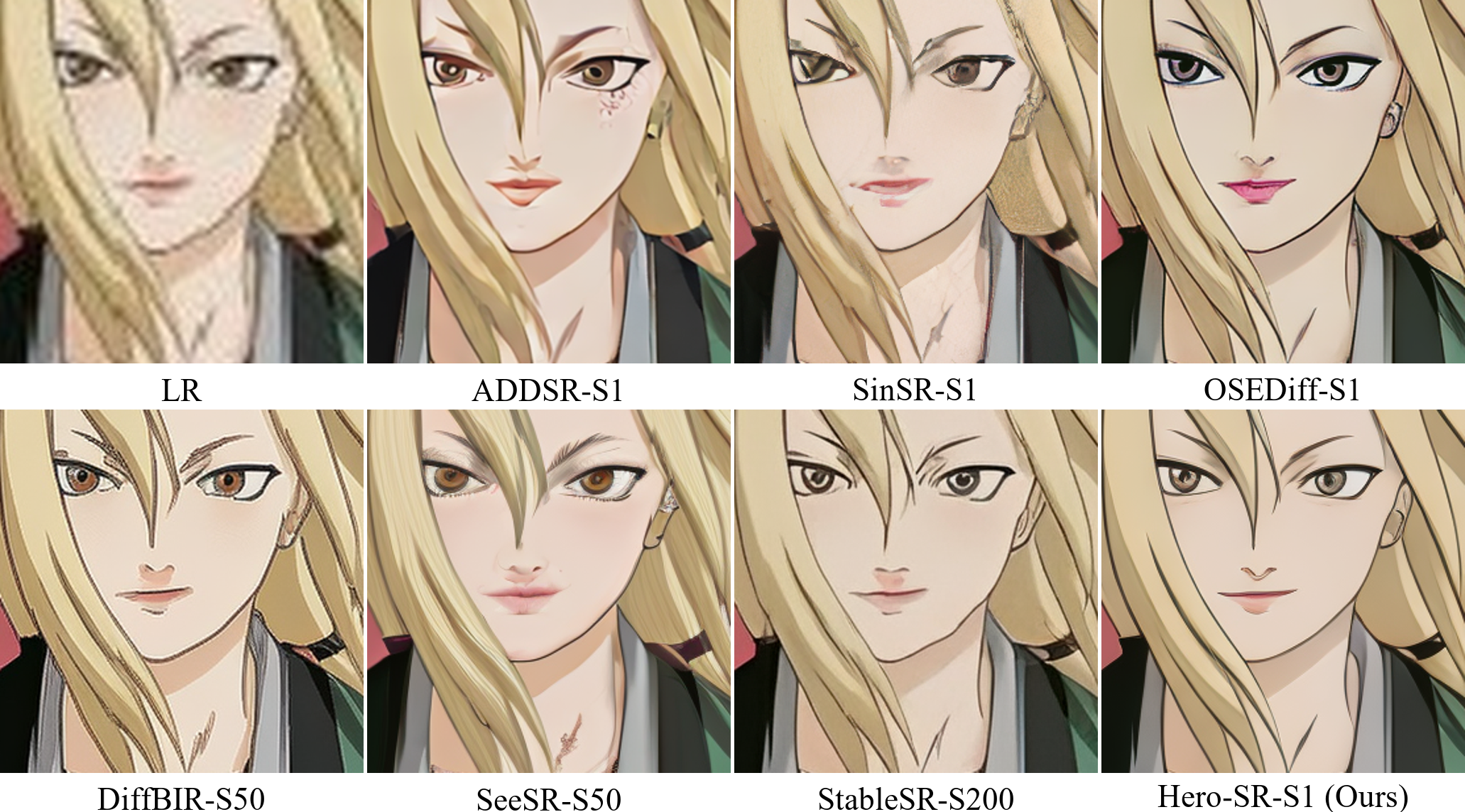}
\caption{Qualitative comparison with one-step and multi-step methods. `S' indicates the number of diffusion steps. \textbf{Zoom in for details.}}
\label{fig:result4}
\end{figure*}

Figures~\ref{fig:result1}, \ref{fig:result2}, \ref{fig:result3}, and \ref{fig:result4} present additional visual comparisons between Hero-SR and other diffusion-based methods. Hero-SR consistently outperforms one-step methods across various scenarios, including architectural structures, animal fur, and text. It also achieves results comparable to or exceeding those of multi-step methods, demonstrating its capability to produce high-quality outputs efficiently. Notably, Hero-SR excels in balancing fine detail restoration and semantic consistency, making its outputs more aligned with human perception across diverse and challenging contexts.

\begin{table*}[ht]
\centering
\small
\caption{Quantitative comparison with \textbf{GAN-base} methods on both synthetic and real-world benchmarks. The best and second best results of each metric are highlighted in {\color[HTML]{FF0000}\textbf{red}} and {\color[HTML]{0000FF}\textbf{blue}}, respectively.}
\resizebox{\linewidth}{!}{
\begin{tabular}{c|c|ccccccccc}
\toprule

DataSet & Methods & PSNR$\uparrow$ & SSIM$\uparrow$ & LPIPS$\downarrow$ & MUSIQ$\uparrow$ & HyperIQA$\uparrow$ & TOPIQ$\uparrow$ & TRES$\uparrow$ & ARNIQA$\uparrow$ & Q-Align$\uparrow$ \\
\midrule

\multirow{5}{*}{DIV2K} 
& BSRGAN & {\color[HTML]{FF0000} \textbf{24.5831}} & 0.6269 & 0.3351 & {\color[HTML]{0000FF} \textbf{61.1928}} & {\color[HTML]{0000FF} \textbf{0.5719}} & {\color[HTML]{0000FF} \textbf{0.5460}} & {\color[HTML]{0000FF} \textbf{74.0277}} & 0.6605 & 2.8535 \\
 & RealESRGAN & 24.2927 & {\color[HTML]{FF0000} \textbf{0.6372}} & {\color[HTML]{0000FF} \textbf{0.3112}} & 61.0570 & 0.5665 & 0.5297 & 70.1277 & {\color[HTML]{0000FF} \textbf{0.6734}} & {\color[HTML]{0000FF} \textbf{3.0684}} \\
 & FeMaSR & 23.0587 & 0.5887 & 0.3126 & 60.8277 & 0.5591 & 0.5231 & 70.7251 & 0.6645 & 2.8828 \\
 & SwinIR & 23.9314 & {\color[HTML]{0000FF} \textbf{0.6285}} & 0.3160 & 60.2177 & 0.5504 & 0.5100 & 68.6045 & 0.6616 & 2.9727 \\
& Hero-SR (Ours)& {\color[HTML]{0000FF} \textbf{24.3663}} & 0.6257 & {\color[HTML]{FF0000} \textbf{0.3111}} & {\color[HTML]{FF0000} \textbf{69.8524}} & {\color[HTML]{FF0000} \textbf{0.6711}} & {\color[HTML]{FF0000} \textbf{0.6948}} & {\color[HTML]{FF0000} \textbf{87.3938}} & {\color[HTML]{FF0000} \textbf{0.7255}} & {\color[HTML]{FF0000} \textbf{3.9968}} \\
\midrule

\multirow{5}{*}{DrealSR} 
 & BSRGAN & {\color[HTML]{0000FF} \textbf{28.7021}} & 0.8028 & 0.2858 & {\color[HTML]{0000FF} \textbf{57.1596}} & {\color[HTML]{0000FF} \textbf{0.5304}} & 0.5060 & {\color[HTML]{0000FF} \textbf{66.7613}} & {\color[HTML]{0000FF} \textbf{0.6262}} & {\color[HTML]{0000FF} \textbf{2.9551}} \\
 & RealESRGAN & 26.8655 & 0.7569 & 0.3157 & 53.7035 & 0.4877 & 0.4673 & 59.3529 & 0.6181 & 2.8711 \\
 & FeMaSR & 28.6147 & {\color[HTML]{FF0000} \textbf{0.8051}} & {\color[HTML]{0000FF} \textbf{0.2819}} & 54.2777 & 0.4938 & 0.4623 & 58.7931 & 0.6101 & 2.8633 \\
 & SwinIR & 28.4969 & {\color[HTML]{0000FF} \textbf{0.8044}} & {\color[HTML]{FF0000} \textbf{0.2743}} & 52.7369 & 0.4800 & 0.4424 & 58.0347 & 0.5948 & 2.8125 \\
\multirow{-5}{*}{DrealSR} & Hero-SR (Ours)& {\color[HTML]{FF0000} \textbf{28.8962}} & 0.8016 & 0.2933 & {\color[HTML]{FF0000} \textbf{66.4874}} & {\color[HTML]{FF0000} \textbf{0.6434}} & {\color[HTML]{FF0000} \textbf{0.6622}} & {\color[HTML]{FF0000} \textbf{83.5888}} & {\color[HTML]{FF0000} \textbf{0.6913}} & {\color[HTML]{FF0000} \textbf{3.6302}} \\
\midrule

\multirow{5}{*}{RealSR} 
 & BSRGAN & {\color[HTML]{FF0000} \textbf{26.3793}} & {\color[HTML]{0000FF} \textbf{0.7651}} & {\color[HTML]{0000FF} \textbf{0.2656}} & {\color[HTML]{0000FF} \textbf{63.2838}} & {\color[HTML]{0000FF} \textbf{0.5617}} & {\color[HTML]{0000FF} \textbf{0.5505}} & {\color[HTML]{0000FF} \textbf{75.7009}} & {\color[HTML]{0000FF} \textbf{0.6830}} &{\color[HTML]{0000FF} \textbf{3.1797}}  \\
 & RealESRGAN & 25.0632 & 0.7356 & 0.2937 & 59.0565 & 0.5215 & 0.5029 & 67.2148 & 0.6674 & 3.0117 \\
 & FeMaSR & 25.6854 & 0.7614 & 0.2709 & 60.3697 & 0.5231 & 0.5148 & 67.6841 & 0.6751 & 3.1055 \\
 & SwinIR & {\color[HTML]{0000FF} \textbf{26.3081}} & {\color[HTML]{FF0000} \textbf{0.7729}} & {\color[HTML]{FF0000} \textbf{0.2539}} & 58.6948 & 0.4973 & 0.4787 & 64.7595 & 0.6609 & 2.9434 \\
\multirow{-5}{*}{RealSR} & Hero-SR (Ours)& 25.8271 & 0.7439 & 0.2893 & {\color[HTML]{FF0000} \textbf{70.0254}} & {\color[HTML]{FF0000} \textbf{0.6623}} & {\color[HTML]{FF0000} \textbf{0.6881}} & {\color[HTML]{FF0000} \textbf{88.5315}} & {\color[HTML]{FF0000} \textbf{0.7170}} & {\color[HTML]{FF0000} \textbf{3.8470}} \\
\bottomrule
\end{tabular}}
\label{tab:gan_result}
\end{table*}

\begin{figure*}[ht]
\centering
\includegraphics[height=0.31\textheight]{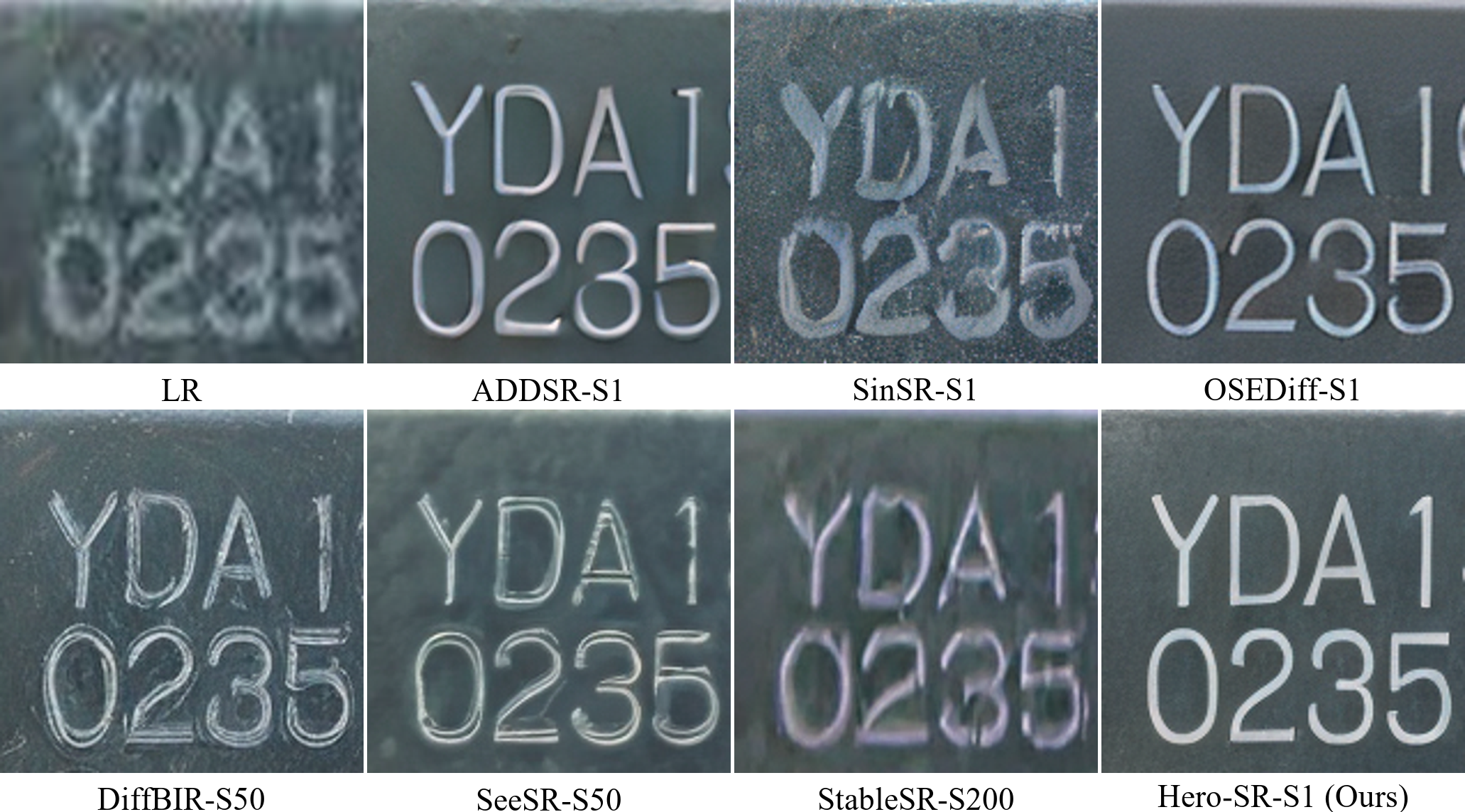}
\end{figure*}

\begin{figure*}[ht]
\centering
\includegraphics[height=0.31\textheight]{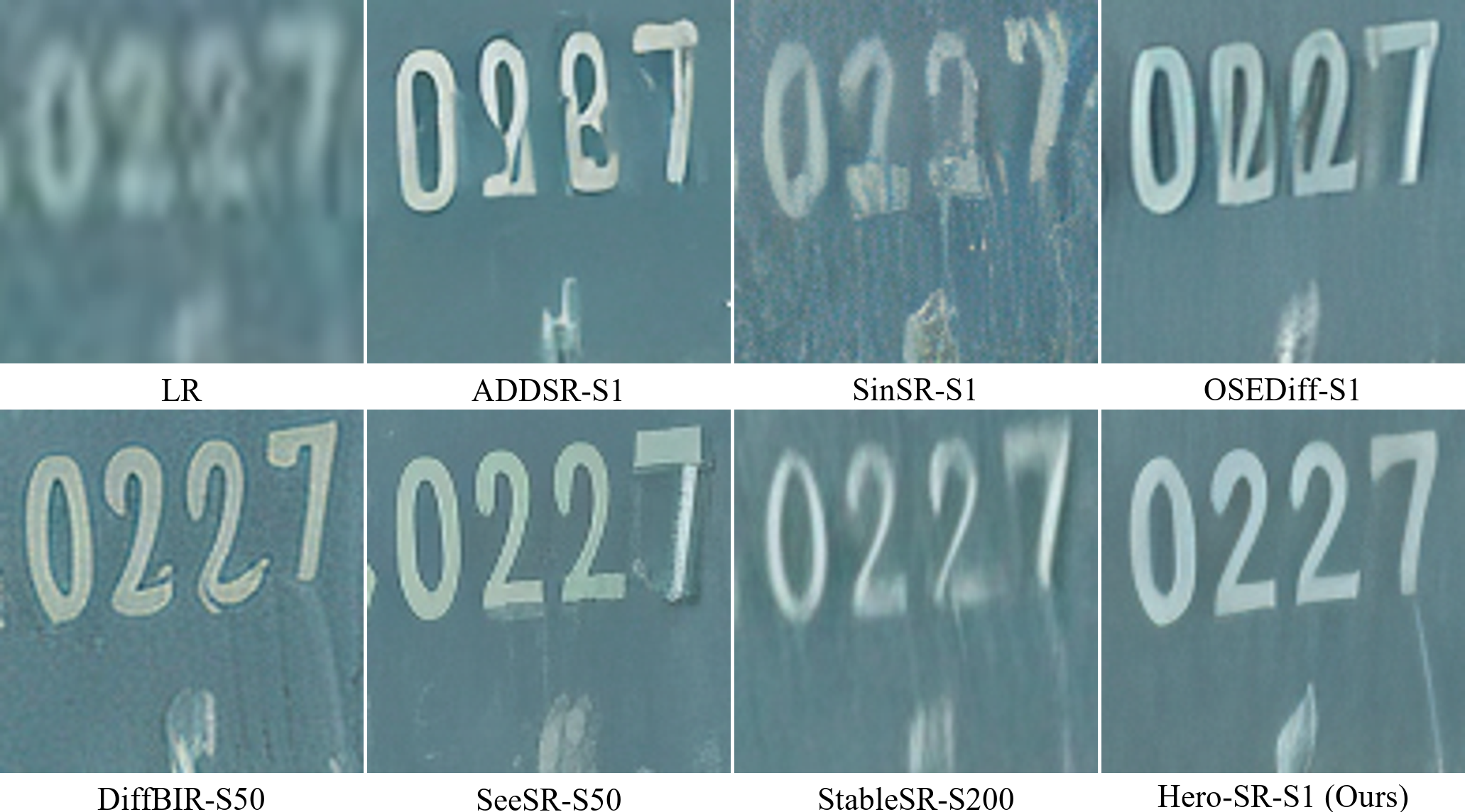}
\end{figure*}

\begin{figure*}[ht]
\centering
\includegraphics[height=0.31\textheight]{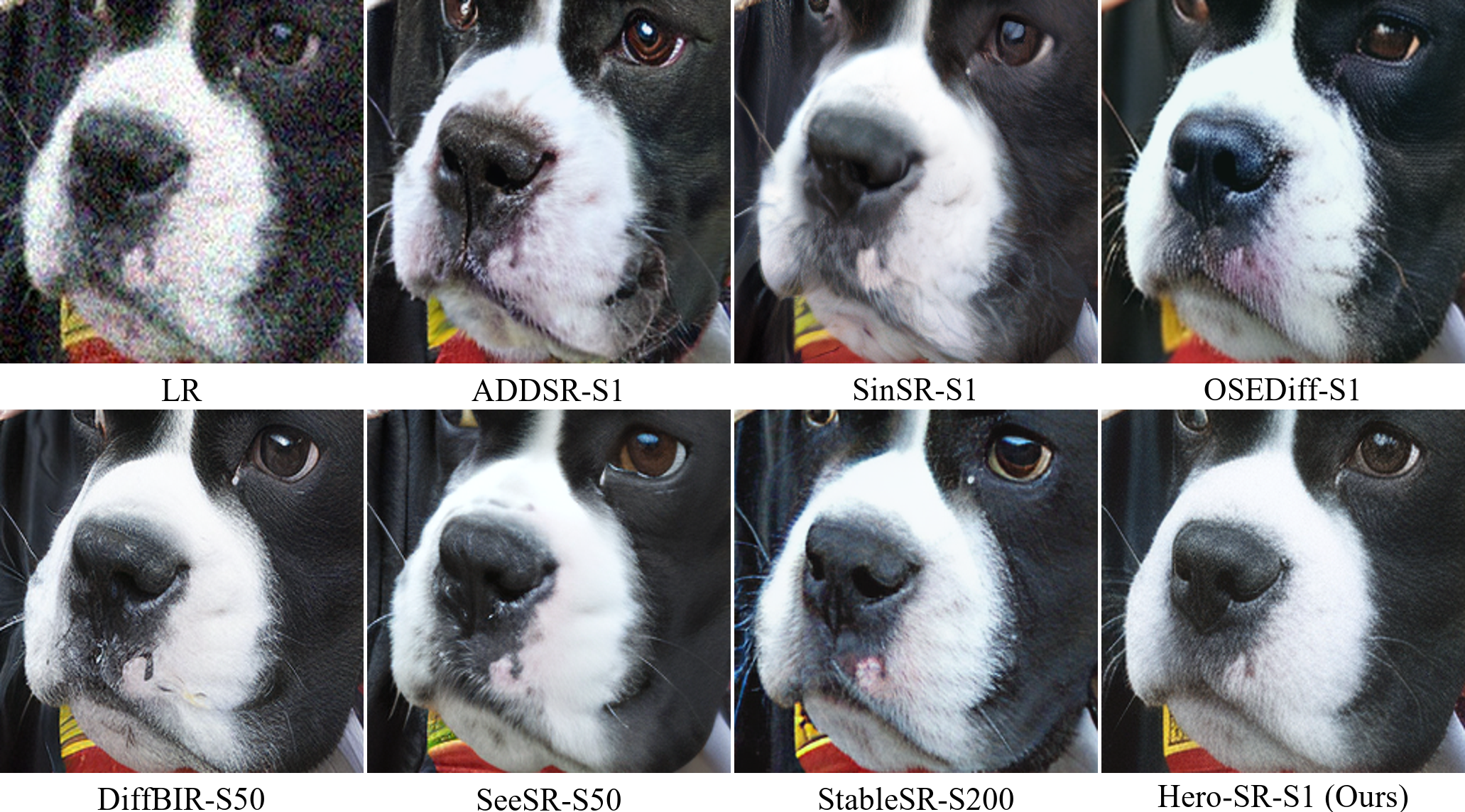}
\end{figure*}

\begin{figure*}[ht]
\centering
\includegraphics[height=0.31\textheight]{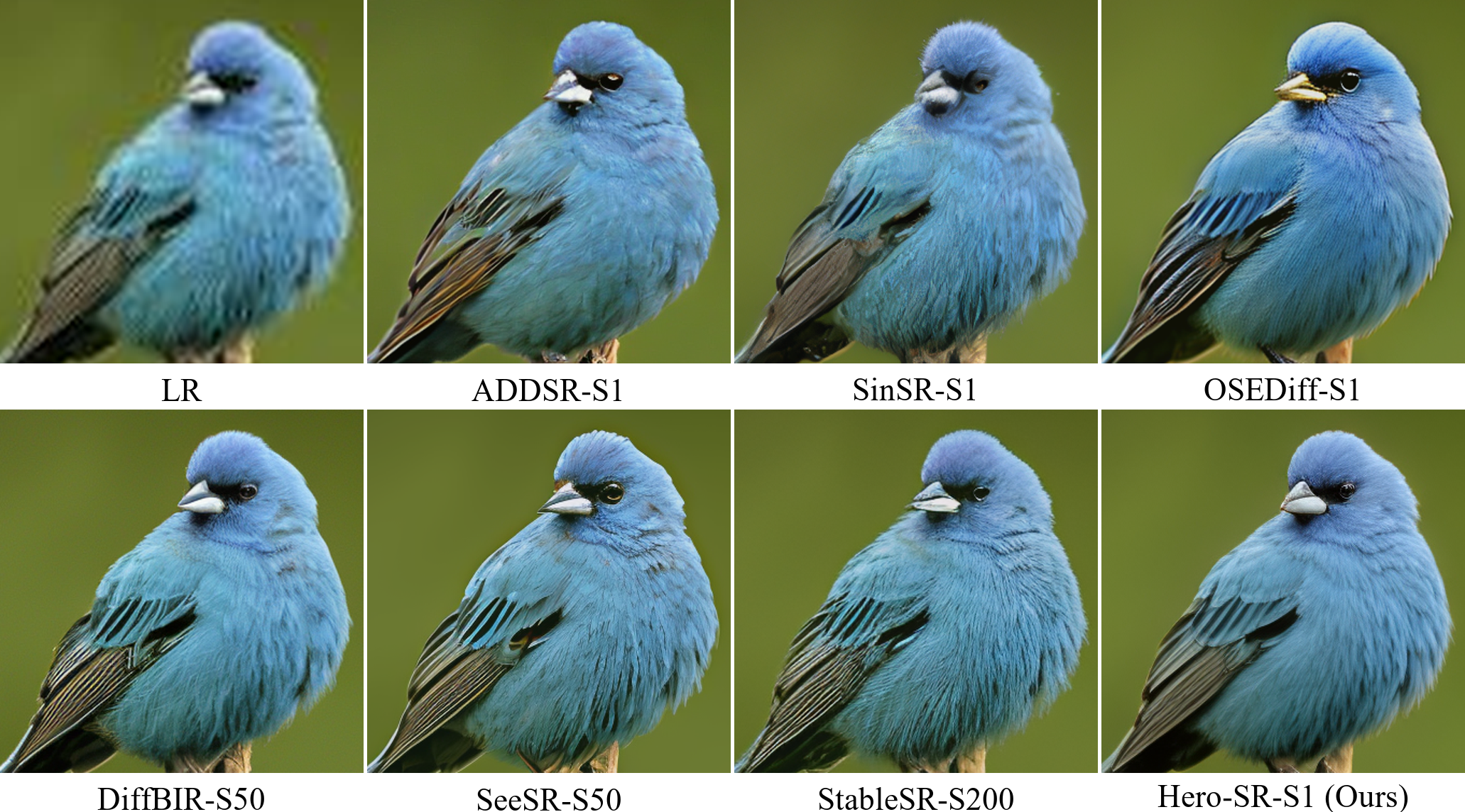}
\end{figure*}

\begin{figure*}[ht]
\centering
\includegraphics[height=0.31\textheight]{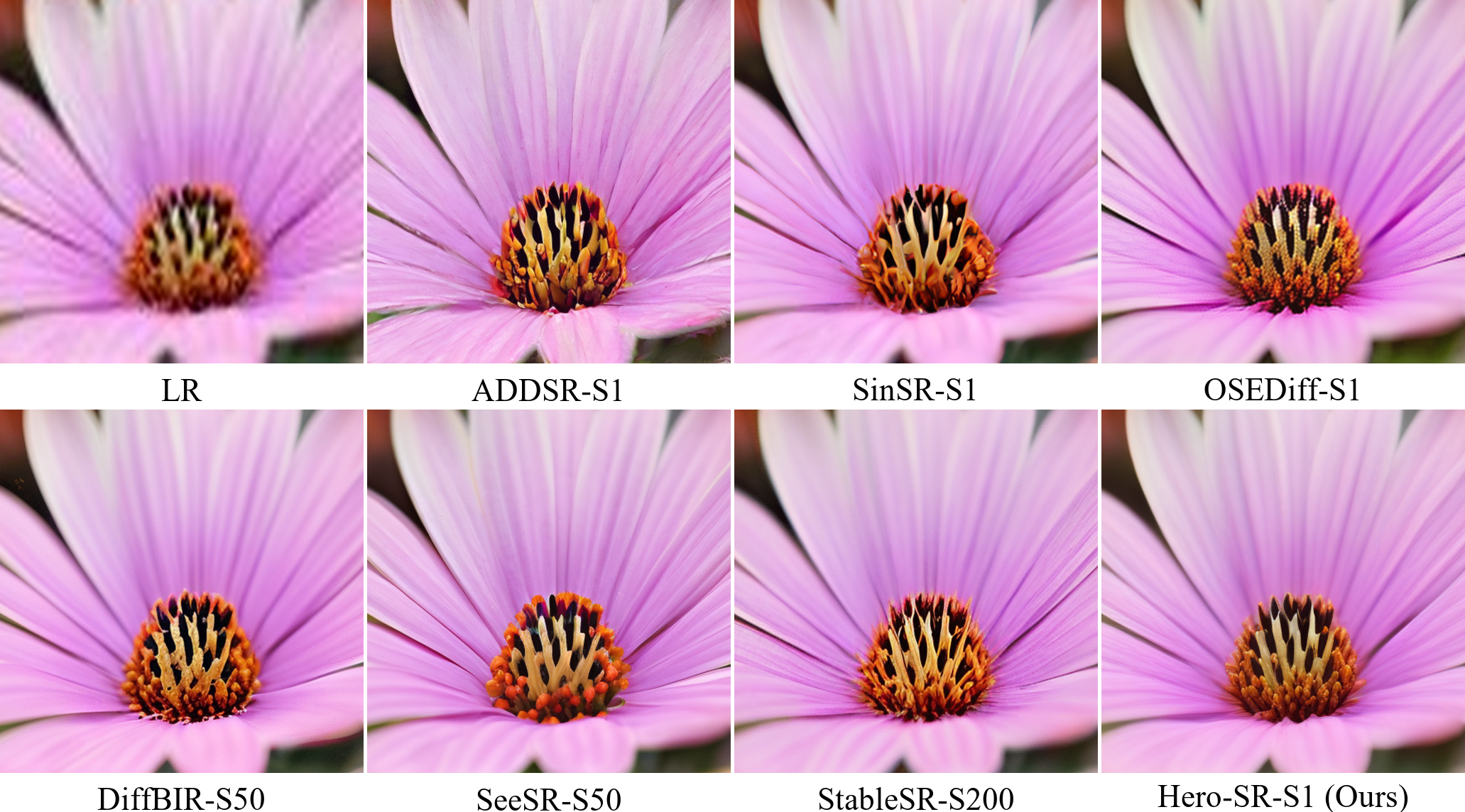}
\end{figure*}

\begin{figure*}[ht]
\centering
\includegraphics[height=0.31\textheight]{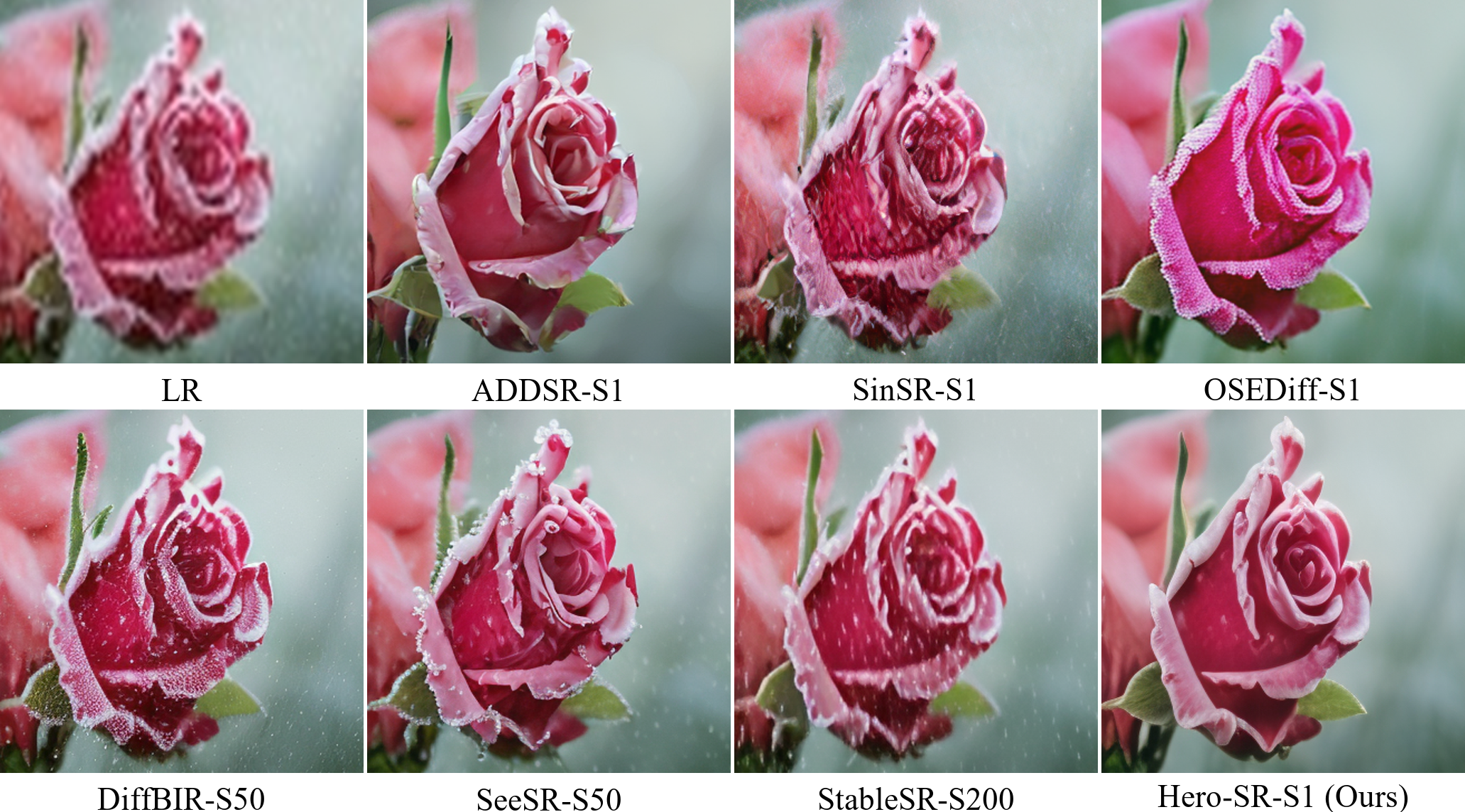}
\end{figure*}

\begin{figure*}[ht]
\centering
\includegraphics[height=0.31\textheight]{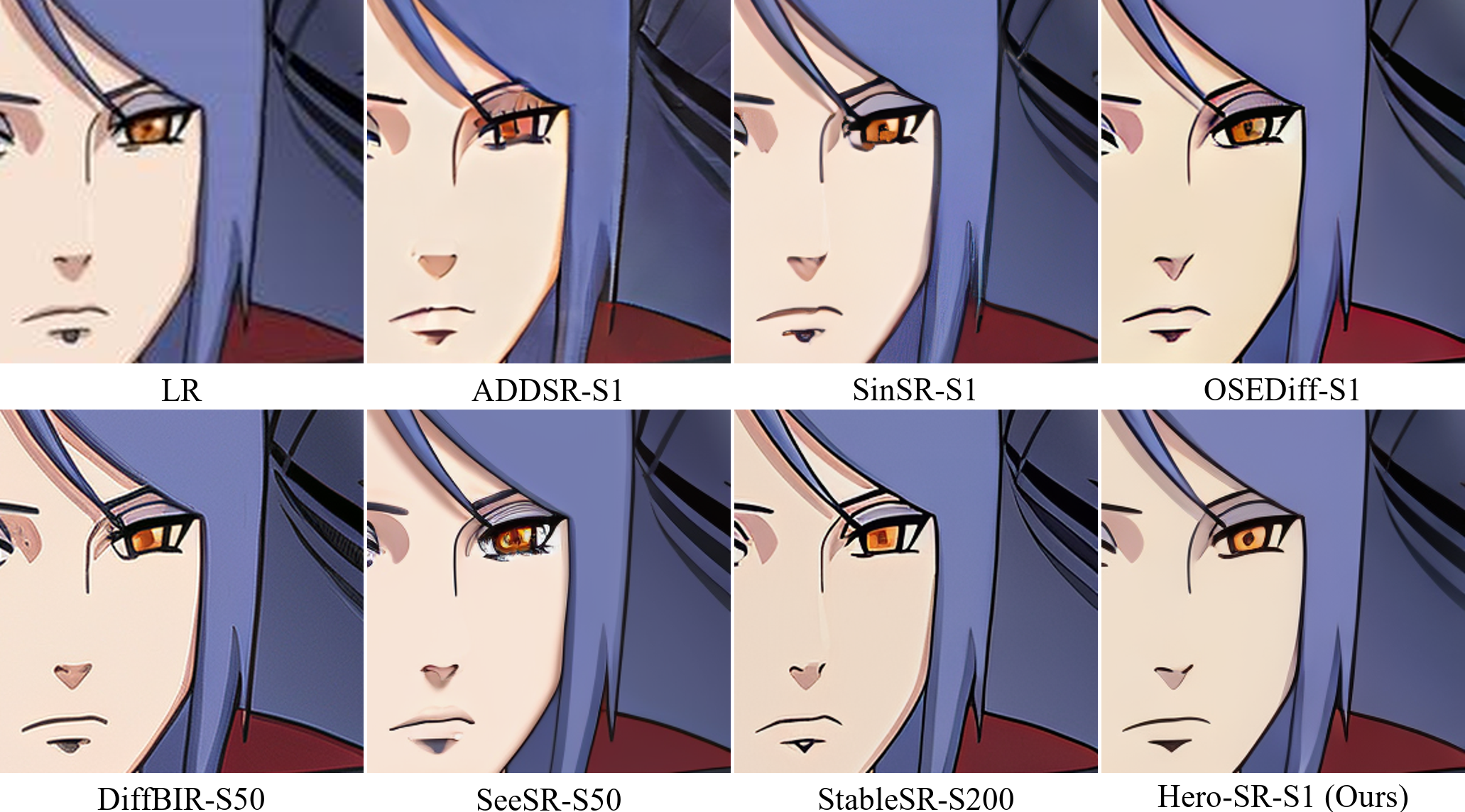}
\end{figure*}

\begin{figure*}[ht]
\centering
\includegraphics[height=0.31\textheight]{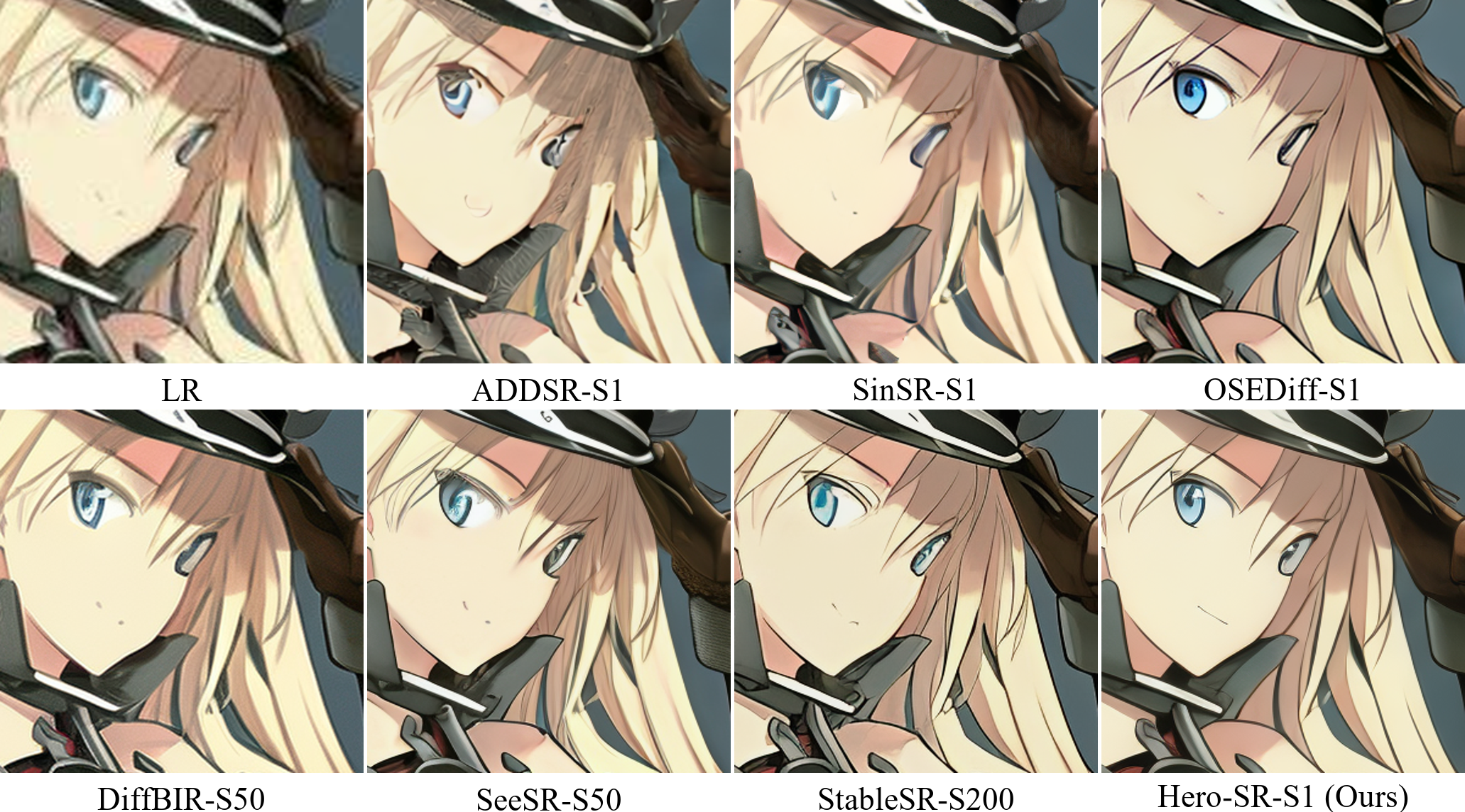}
\end{figure*}

\end{document}